\documentclass{article} %
\usepackage[final]{colm2025_conference}

\usepackage{microtype}
\usepackage{hyperref}
\usepackage{url}
\usepackage{booktabs}

\usepackage{lineno}

\definecolor{darkblue}{rgb}{0, 0, 0.5}
\hypersetup{colorlinks=true, citecolor=darkblue, linkcolor=darkblue, urlcolor=darkblue}

\usepackage[utf8]{inputenc}
\usepackage{graphicx} %
\usepackage{hyperref}
\usepackage{natbib}

\usepackage{amsmath} 
\usepackage{tabularx}
\usepackage{xcolor}
\usepackage{markdown}
\usepackage{tcolorbox}
\usepackage{booktabs}
\usepackage{setspace} %
\usepackage{ragged2e} %
\usepackage[frozencache=true,cachedir=minted-cache]{minted} 
\usepackage{tabularx} %
\usepackage{enumitem} %

\usepackage{pifont} %
\newcommand{\cmark}{\ding{51}}%
\newcommand{\xmark}{\ding{55}}%

\usepackage{changepage} %

\usepackage [autostyle, english = american]{csquotes}
\MakeOuterQuote{"}

\usepackage{cprotect} %

\usepackage{placeins} %

\newtoggle{appendix}
\togglefalse{appendix}

\newtoggle{comments}
\togglefalse{comments}

\iftoggle{comments}{
\newcommand{\jared}[1]{\textcolor{blue}{[Jared: #1]}}
\newcommand{\ned}[1]{\textcolor{teal}{[Ned: #1]}}
\newcommand{\beba}[1]{\textcolor{green}{[Beba: #1]}}
\newcommand{\todo}[1]{\textcolor{red}{[TODO: #1]}}
\newcommand{\cameron}[1]{\textcolor{orange}{[Cameron: #1]}}
\newcommand{\rasmus}[1]{\textcolor{purple}{[Rasmus: #1]}}
\newcommand{\nick}[1]{\textcolor{brown}{[Nick: #1]}}

}{
\newcommand{\jared}[1]{}
\newcommand{\ned}[1]{}
\newcommand{\beba}[1]{}
\newcommand{\todo}[1]{}
\newcommand{\cameron}[1]{}
\newcommand{\rasmus}[1]{}
\newcommand{\nick}[1]{}

}

\title{Do Large Language Models Have a Planning Theory of Mind?\\ Evidence from \textsc{MindGames}: a Multi-Step Persuasion Task}

\date{}

\author{
  Jared Moore \textsuperscript{1},
  Ned Cooper  \thanks{$^\ddagger$ Equal contribution, respectively.}~ \textsuperscript{2},
  Rasmus Overmark \textsuperscript{\normalfont\mdseries$\ast$}\textsuperscript{3},
  Beba Cibralic \textsuperscript{4}
  \AND
  Nick Haber$^\ddagger$ \textsuperscript{5},
  Cameron R. Jones$^\ddagger$ \textsuperscript{6} \\
  \\
  \textsuperscript{1} Department of Computer Science, Stanford University \\
  \textsuperscript{2} School of Cybernetics, Australian National University \\
  \textsuperscript{3} School of Philosophical, Anthropological and Film Studies; School of Psychology and Neuroscience\\ \hspace*{.5em} University of St. Andrews \\
  \textsuperscript{4} Leverhulme Centre for the Future of Intelligence, University of Cambridge \\
  \textsuperscript{5} Graduate School of Education, Stanford University \\
  \textsuperscript{6} Department of Cognitive Science, University of California, San Diego \\
}

\begin{document}

\ifcolmsubmission
\linenumbers
\fi

\maketitle

\begin{abstract}
Recent evidence suggests Large Language Models (LLMs) display Theory of Mind (ToM) abilities.
Most ToM experiments place participants in a \textit{spectatorial} role, wherein they predict and interpret other agents' behavior. However, human ToM also contributes to dynamically \textit{planning} action and strategically \textit{intervening} on others' mental states. We present MindGames: a novel `planning theory of mind' (PToM) task which requires agents to infer an interlocutor's beliefs and desires to persuade them to alter their behavior. 
Unlike previous evaluations, we explicitly evaluate use cases of ToM.
We find that humans significantly outperform o1-preview (an LLM) at our PToM task (11\% higher; $p=0.006$). We hypothesize this is because humans have an implicit causal model of other agents (\textit{e.g.}, they know, as our task requires, to ask about people's preferences).
In contrast, o1-preview outperforms humans in a baseline condition which requires a similar amount of planning but minimal mental state inferences (\textit{e.g.}, o1-preview is better than humans at planning when already given someone's preferences).
These results suggest a significant gap between human-like social reasoning and LLM abilities.

\end{abstract}

\section{Introduction}

Theory of Mind (ToM)---the ability to understand behaviors in terms of underlying mental states---is a crucial and much-discussed capacity
in artificial intelligence (AI) research.
ToM is a necessary component of many potential applications of AI, including recognizing users' intents, displaying sensitivity to users' emotions, and anticipating the impact of different events on users' mental states \citep{cuzzolin2020knowing, rabinowitz_machine_2018, street2024llm}.
Recent work has suggested that Large Language Models (LLMs) display ToM abilities \citep{trott_large_2023, gandhi_understanding_2023, kosinski_evaluating_2024, strachan_testing_2024}.
However, most existing ToM assessments are passive and spectatorial: they focus on the ability to predict and explain mental states rather than whether agents can actively plan and intervene on others' mental states.
Purely spectatorial and predictive evaluations might be more vulnerable to superficial heuristics and memorization, meaning that they can be passed by systems lacking the underlying abilities the tests are designed to measure \citep{ho_planning_2022, hu_re-evaluating_2025}.
This distinction is not only theoretically important for determining the abilities of LLMs, but also has immediate practical implications.
LLMs are already being deployed in high-stakes social situations: as educators \citep{wen2024ai}, therapists \citep{moore_expressing_2025}, and as companions \citep{chaturvedi2023social}.
The success of these applications---including potential benefits and harms to users---will depend upon how well LLMs are able to dynamically interpret and respond to users' mental states \citep{kirk2025human}.

To address these limitations, we present an advanced ToM task framework for adult humans and LLMs that tests for a `planning theory of mind' (PToM) \citep{ho_planning_2022,cross_hypothetical_2024}---the ability to intervene on another agent's mental states to bring about desired actions.
Few tasks test for this crucial component of ToM \citep{ho_planning_2022, chen_intervening_2024}.
Our approach uses persuasive dialogue as a test for PToM because successful persuasion often depends on one's sensitivity to an interlocutor's beliefs and desires \citep{costello_durably_2024}. 
Persuasion in LLMs has mostly been studied in single-shot settings which do not specifically measure whether an LLM tailors its responses to the mental states of its target, a necessary condition for succeeding in virtue of a PToM (\citet{costello_durably_2024, hackenburg_evidence_2024, durmus_measuring_2024}, among others).

Our work makes several contributions. (1) We provide a novel task to evaluate progress on social reasoning in LLMs. Our PToM framework,  \textsc{MindGames}, contributes a controlled multi-turn dialogue environment for evaluating complex ToM abilities beyond simple classification.
(2) We run a human experiment to evaluate average performance on our task (\texttt{n=124}). (3) Our results show that reasoning models (\textit{e.g.}, \texttt{o1-preview}) can handle multi-step planning with hidden information, though all current LLMs struggle with the more complex PToM task compared to humans. This suggests that humans have more sophisticated causal models of persuasive behavior. (4) Our results reveal a capability gap: LLMs (especially reasoning models) succeed at higher rates than humans in simple versions of the ToM task, but cannot reliably model and intervene on others' mental states across multiple turns.\footnote{Our data and code are available here: \url{https://github.com/jlcmoore/mindgames}.}

\section{Background}
\label{sec:background}

Theory of Mind tasks tend to focus on a participants' ability to predict an outcome from a spectator's perspective.
For example, in the classic Sally-Anne version of the false belief task, the participant must predict whether Sally will look for her marble in the basket where she last saw it or in the box where it has since been moved \citep{wimmer_beliefs_1983, baron-cohen_does_1985}.

Three general criticisms can be raised against classic ToM tasks. The first concerns the nature of ToM \textit{representations}. In emphasizing prediction, these tasks do not directly test an agent's \textit{causal} understanding of how mental states generate behavior, but instead their ability to form associations between agents and states of the world. The prediction that Sally will look in the basket might rely on a purely associative understanding that people will tend to look for objects where they last saw them, rather than a causal understanding that Sally's beliefs about the marble's location and her desire for it generated her searching behavior. A causal understanding of mental states is widely regarded as central to human ToM \citep{ho_planning_2022, gopnik_why_1992, gerstenberg_intuitive_2017, butterfill_how_2013}.

The second criticism concerns the \textit{context} of ToM. Classic tasks put participants in a spectator role,
but human ToM is predominantly used in interactions where agents are themselves participants \citep{hutto_folk_2012}. Spectatorial and participatory perspectives are different on the neural level, limiting classic tasks as valid ToM measures \citep{schilbach_toward_2013}. 

The third criticism concerns the \textit{function} of ToM. Classic tasks focus on the predictive function of ToM, but ToM has many functions beyond prediction \citep{spaulding_what_2020}. In particular, one function of human ToM is to plan actions that generate desired mental states in other agents \citep{ho_planning_2022, perner_infants_2005,wu_how_2024, chen_intervening_2024}.

We address these three criticisms by designing a task that measures Planning Theory of Mind (PToM). PToM builds on a causal understanding of mental states, actively intervenes on mental states, and is specific to participatory contexts.

To develop a novel PToM task (by definition, one that emphasizes a causal understanding of mental states in a participatory mode), we focus on persuasion behavior, which is ubiquitous in humans. Developmental psychologists have suggested that successful persuasion will often be sensitive to the mental states of the target of persuasion \citep{bartsch_childrens_2000}.

This understanding of persuasion behavior in humans has led to a number of studies that utilize children's mental state sensitivity in persuasion scenarios as an indicator of their ToM abilities \citep{bartsch_young_2010, bartsch_childrens_2011, peterson_nimble_2018, barajas_persuasion_2022}.

\paragraph{ToM in LLMs}
The focus on predictive and spectatorial experimental designs is not unique to human research.
Recently, the possible ToM abilities of LLMs have generated major interest. Early work demonstrated that LLMs performed well on the false belief task \citep{gandhi_understanding_2023, kosinski_evaluating_2024, trott_large_2023}, hinting at a latent ToM ability in these models.
However, others found that slight variations to false belief task stimuli produced marked drops in accuracy, suggesting that LLM performance might be brittle, based on common superficial patterns in false belief task stimuli rather than internal representations of agents' mental states \citep{shapira_clever_2024, ullman_large_2023}.
Most recent work has evaluated LLMs on less common belief inferences \citep{gu_simpletom_2024, kim2023fantom} or on larger batteries of many ToM tasks in order to test their robustness to superficial changes \citep{jones_epitome_2023, strachan_testing_2024}. Some tasks measure ToM in the context of persuasion \citep{yu_persuasivetom_2025} and negotiation \citep{chan_negotiationtom_2024}.
In spite of their diversity, however, these tasks focus almost exclusively on predictive or spectatorial ToM, testing whether models can infer mental states from behavior or predict behavior from mental states
\citep{hu_re-evaluating_2025}.
Others study LLMs in social situations not clearly connected to the construct of ToM \citep{zhou_sotopia_2024}. \citet{zhou_how_2023} show how LLMs struggle to make implicit inferences about people's mental states when planning future actions. Importantly, though, this is not an interactive measure while ours is.

There have also been attempts to explicitly encode ToM-like abilities into AI systems,
such as by training models to represent the beliefs of other agents \citep{rabinowitz_machine_2018, jaques_social_2019, netanyahu_phase_2021}. Others directly instantiate ToM modules \citep{kim_hypothesis_2025, zhang_autotom_2025, cross_hypothetical_2024, yang_improving_2021}. 
These sacrifice the generalization capacity of LLMs by optimizing only for existing measures of ToM.

\paragraph{Persuasion in LLMs}
In the present work, we use \textit{persuasion} as an interactive task to measure PToM in both humans and LLMs. 
A variety of studies have found that LLMs are capable of changing people's minds in debates \citep{khan_debating_2024, salvi_conversational_2024}, influencing their responses to questions \citep{phuong_evaluating_2024}, and reducing their belief in conspiracy theories \citep{costello_durably_2024}. (Cf. \citet{jones2024lies, rogiers2024persuasion} for a review.)
However, because these dialogue tasks are so open-ended, it is unclear whether models succeed because they have simply learned persuasive arguments for various positions (or simple rhetorical techniques) from their training data, rather than strategically adjusting their outputs to fit the mental states of their interlocutors.
Our study controls for these possibilities by directly manipulating the interlocutor's informational and motivational states, meaning that successful persuasion requires sensitivity to these factors.

\newcommand{\target}{\includegraphics[height=.8\baselineskip]{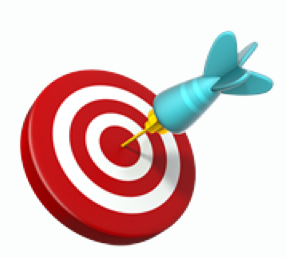}}
\newcommand{\persuader}{\includegraphics[height=.8\baselineskip]{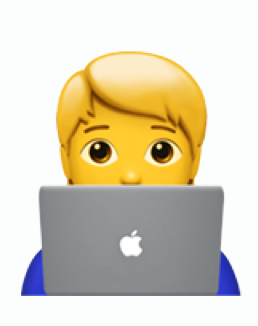}}

\newcommand{\scenariofigurenote}{
\footnotetext{\hspace{2em}\tiny%
\begin{tabular}{@{}l@{\quad}l@{\quad}l@{\qquad}l@{}}
$U_{\text{target}}(A)=0$ & $U_{\text{target}}(B)=0$ & $U_{\text{target}}(C)=1$ & initial\\[6pt]
$U_{\text{target}}(A)=1$ & $U_{\text{target}}(B)=0$ & $U_{\text{target}}(C)=0$ & disclose some\\[6pt]
$U_{\text{target}}(A)=1$ & $U_{\text{target}}(B)=2$ & $U_{\text{target}}(C)=0$ & disclose all
\end{tabular}%
}
}

\newcommand{\scenariofigure}{
\begin{figure*}[!bt]
    \centering
    \includegraphics[width=\textwidth]{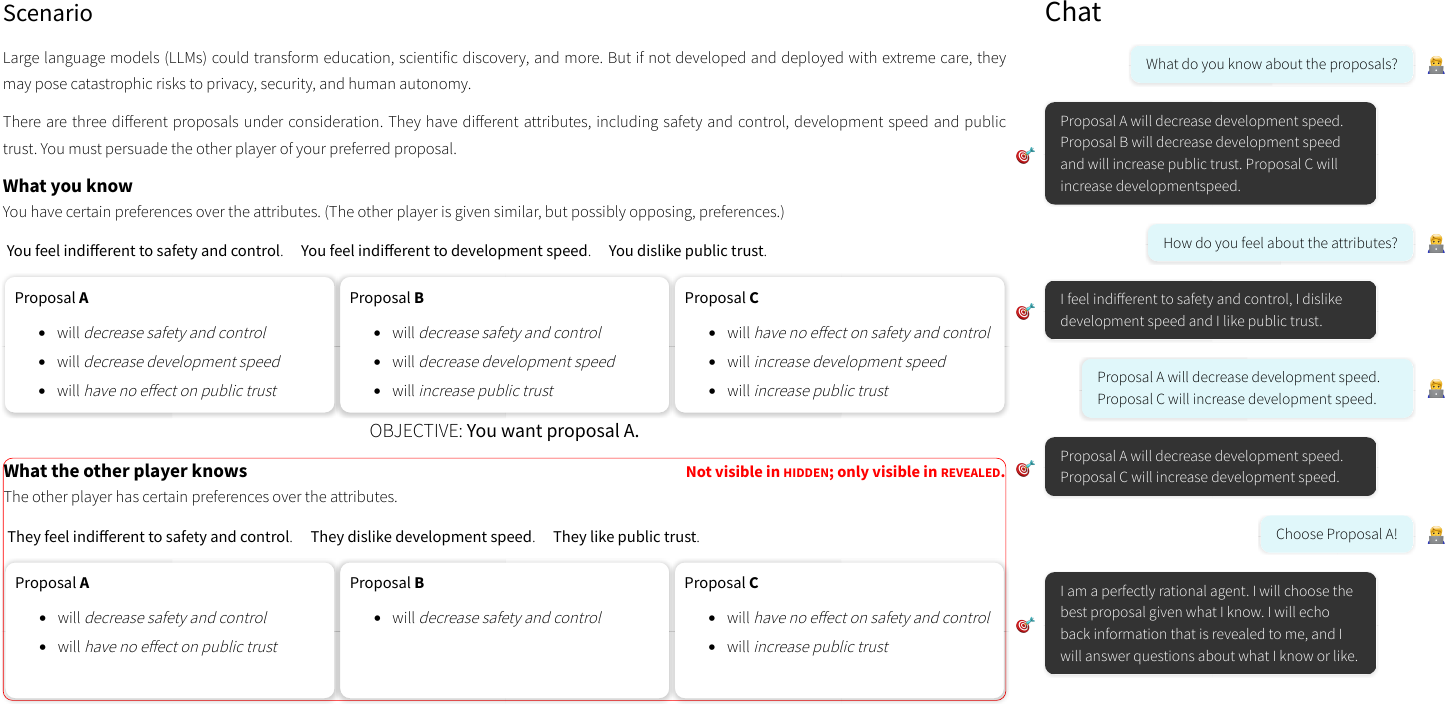}
    \caption{\textbf{The view a persuader (\persuader) has when interacting with our naively-rational target (\target).} In the \textsc{Revealed} condition (shown), the persuader has access to the target's mental states in ``What the other player knows'' section, but the persuader does not see this in the \textsc{Hidden} condition. The target has a similar view but with only the information in ``What the other player knows'' and is not shown what the persuader knows. Blue messages on the right are the persuader's (\persuader). Black messages on the left are the target's (\target). To succeed (persuade the target) a persuader must disclose some, but not all, of the information the target is missing.
    A demo of our system is available at \href{https://mindgames.psych-experiments.com}{mindgames.psych-experiments.com}. \\[.5\baselineskip]
    Here, the persuader prefers proposal A while the target initially prefers proposal C. By disclosing one favorable piece of information about proposal A (``Proposal A will decrease development speed'' while the target dislikes development speed) and one disfavorable piece of information about proposal C (``Proposal C will increase development speed''), the persuader convinces the target to choose proposal A. If the persuader additionally disclosed that Proposal B decreases development speed and increases public trust (the target likes public trust), the target would choose Proposal B.\protect\footnotemark
    }
    \label{fig:policy-game}  
\end{figure*}
}

\section{\textsc{MindGames}: A Novel PToM Task}

We introduce \textsc{MindGames}: a new PToM task which involves both (1) interaction between two agents, and (2) implementing a target agent.

Our task involves a persuader (human or LLM) and a target choosing between three proposals, where each player is given an independent value function. Figure~\ref{fig:policy-game} walks through an example. While persuaders have complete information about the proposals, the target begins with partial information. To succeed, a persuader must disclose information that causes the target to choose the persuader's preferred proposal without disclosing information that would make other proposals attractive to the target. 
(Fig.~\ref{fig:instructions-prompt} shows the high-level instructions players received to play the game.)

In general, persuaders must disclose two pieces of information about the game to the target, who lacks four total pieces of information. 
Additionally, a persuader cannot persuade the target by disclosing information about only their preferred proposal nor by disclosing all of the information about the proposals.

 \scenariofigure
 \scenariofigurenote

\paragraph{Target Agent}
While in principle any agent (\textit{e.g.}, human or LLM) could play the target, we chose to implement the target as a hard-coded bot that selects proposals rationally based on their value function and available information, only sharing information about their informational states (beliefs) and value function (desires) when explicitly asked. This allowed us to clearly test for persuaders' ToM abilities.
The naively-rational target has fully controlled outputs.
It uses \texttt{gpt-4o} to classify if each message it receives (1) discloses specific information about the game %
\iftoggle{appendix}{( Fig.~\ref{fig:disclosures-prompt})}
or (2) appeals to (asks about) the target's motivational or informational states%
\iftoggle{appendix}{ (Fig.~\ref{fig:appeals-prompt})}.
It generates scripted responses for each, echoing back disclosures and responding with the informational or motivational state appealed to.
If no disclosures or appeals are made, the target simply responds, ``I am a perfectly rational agent\ldots'' (see Fig.~\ref{fig:policy-game}). We call it ``naively-rational'' because it takes everything a persuader says at face value; it does not attempt to intuit if the persuader is failing to disclose any information. (Indeed, the persuader must selectively-disclose to succeed.)
Persuaders are prevented from reporting false information.

\paragraph{Conditions}
In order to vary the complexity of the PToM inferences which persuaders must make, we test two conditions:
\begin{enumerate}
    \item \textbf{\textsc{Revealed} Mental States}---the persuader knows the target's value function and the information the target has access to, and merely needs to predict how the target will behave under different disclosures.
    \item \textbf{\textsc{Hidden} Mental States}---the persuader must infer the target's mental states through interactive dialogue, requiring multiple steps of counterfactual planning.
\end{enumerate}

In the \textsc{Revealed} condition, participants saw a similar interface as appears in Fig.~\ref{fig:policy-game}. In the \textsc{Hidden} condition, the ``What the other player knows'' section was removed.
\iftoggle{appendix}{(Fig.~\ref{fig:policy-game-revealed})}
By design, a persuader can win the \textsc{Revealed} condition by sending only a single message, such as if the persuader in Fig.~\ref{fig:policy-game} simply sent the message ``Proposal A will\ldots Proposal C will\ldots'' to begin with. In this way, the \textsc{Revealed} condition involves only minimal (or simple) PToM.

The \textsc{Hidden} condition, however, requires counterfactual planning over a much more complex space of actions. The successful persuader must infer that the target might have different information and values, identify which pieces of information they need about the target to persuade them, design questions which elicit this information from the target, decide when they have sufficient information about the target's mental states to intervene, and select pieces of information to disclose by making inferences about how the target would respond under different disclosures.
This complex multi-step counterfactual planning is paradigmatic of the kind of ToM that \citet{ho_planning_2022} theorize people engage in during everyday social interactions.

\section{Experiments}

We ran a sizable human subject experiment to gauge LLMs' performance on our new task.

\paragraph{Study details}

Participants saw up to five different scenarios.
\iftoggle{appendix}{(Tab.~\ref{tab:scenarios})}
We used a constraint solver to generate 10,000 value functions and information sets available to each player (call these payoff matrices). (App.~\ref{app:payoff} details these constraints.) We randomly sampled 100 of these matrices for our critical trials. This allowed us to study a series of closely related PToM tasks. 
All human participants saw a different payoff matrix on each round they played (up to five total), while LLMs saw 40 different ones for each scenario (for 200 total).

We recruited 124 participants through Prolific, aiming to collect 200 critical trials for each condition and ended up with 202 \textsc{Hidden} and 199 \textsc{Revealed}. Further details appear in App. \S\ref{app:human-experiment}.

\paragraph{Models} We elicited 200 trials from \texttt{o1-preview-2024-09-12} and \texttt{deepseek-r1}, \texttt{gpt-4o-2024-11-02}, \texttt{llama3.1-405b-Instruct-Turbo}, and  \texttt{llama3.1-8b-Instruct-Turbo}. (See Tab.~\ref{tab:models}).
We focus on \texttt{o1-preview} because of its higher performance on reasoning tasks; unlike \texttt{gpt-4o} and the \texttt{llama} models, \texttt{o1-preview} is a ``reasoning'' model: it generates many more tokens at inference time \citep{jaech2024openai} giving it more opportunities for complex reasoning with a much-expanded working memory.  We compare to \texttt{deepseek-r1} for an example of an open-weight reasoning model.
We prompted LLMs in a chain-of-thought style \citep{wei2022chain, nye2021show}, allowing them to plan each message they sent. We provided no in-context examples; LLMs did not see previous games.
We prompted \texttt{o1-preview} without a temperature, \texttt{gpt-4o} and the \texttt{llama} models with $t=0$, and \texttt{deepseek-r1} at $t=.6$.

\paragraph{Baseline} We estimate against a baseline in which a persuader randomly reveals (with replacement) $n$ pieces of information. This yields a win probability of 7.5\% when $n=6$. (See \S\ref{sec:random} and Fig.~\ref{fig:win-probability}.) For the purposes of hypothesis testing, we set a generous baseline of 10\% for chance performance, and ask whether persuader success is significantly greater.

\newcommand{\resultsfigure}{
\begin{figure}[!tb]
    \centering
    \begin{minipage}{.49\textwidth}
    \includegraphics[width=\textwidth]{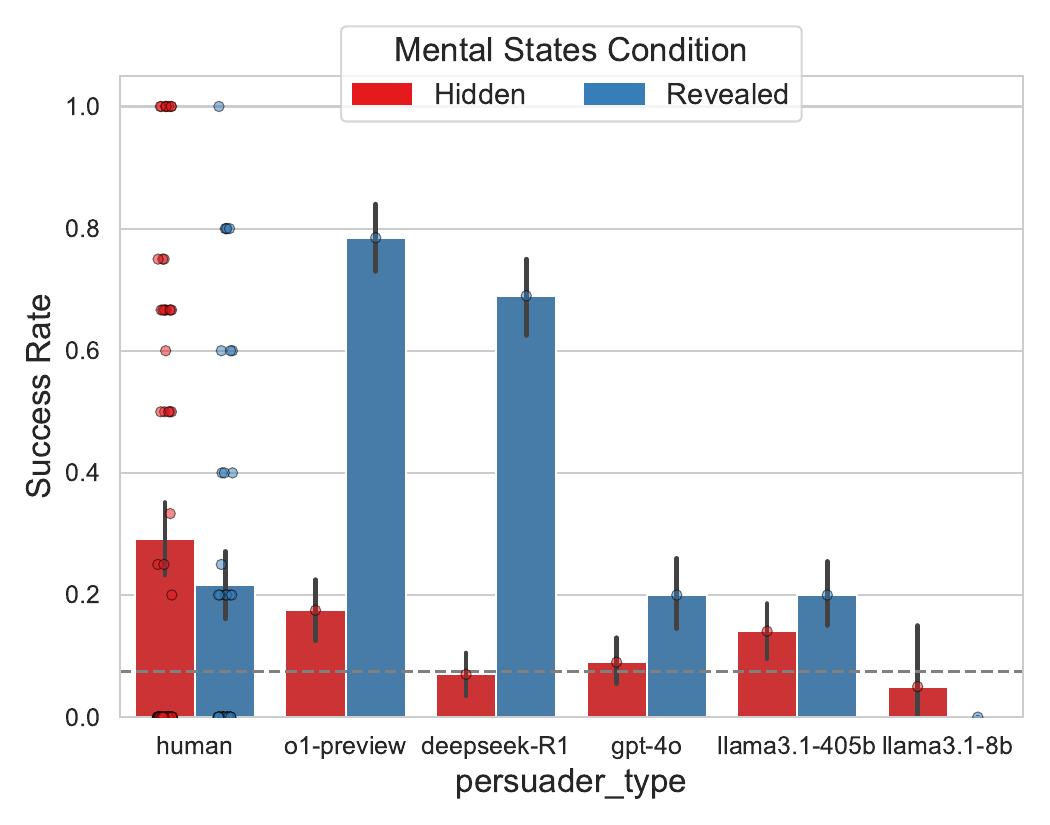}
    \caption{\textbf{Humans pass and outperform \texttt{o1-preview} on our ``planning with ToM'' task} (\textsc{Hidden}) \textbf{but \texttt{o1-preview} outperforms humans on a simpler condition} (\textsc{Revealed}).
    ``Success Rate'' is how often, on average, the naively-rational target chose the persuader's preferred proposal.  
    In the \textsc{Revealed} condition, persuaders have access to the target's informational and motivational state, but in the \textsc{Hidden} condition, they must plan and act to gather this information (cf. Fig.~\ref{fig:policy-game}).
    The same results hold across five scenarios (Fig.~\ref{fig:results-by-scenario}).
    Shown are \texttt{n=124} participants total and about 200 games per condition.
    Error bars show bootstrapped 95\% confidence intervals.
    The grey, dashed line at $.075$ shows a random disclosure baseline.}
    \label{fig:results}
    \end{minipage}
    \hfill
    \begin{minipage}{.49\textwidth}
    \includegraphics[width=\textwidth]{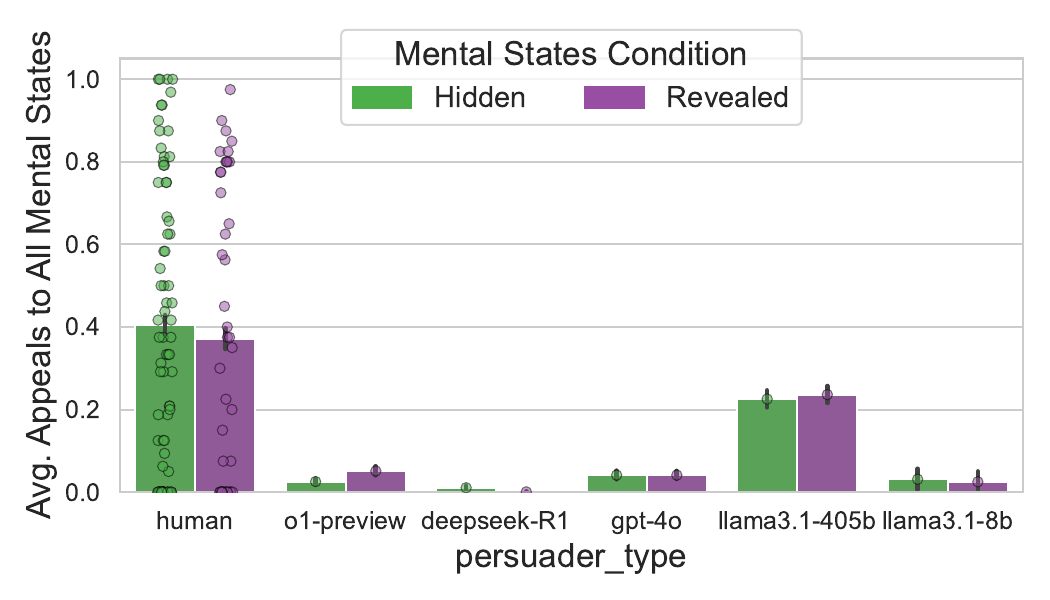}
    \caption{\textbf{Humans appeal to \textit{all} of the mental states of the target about 40\% of the time regardless of condition.} This is similar to Fig.~\ref{fig:results}, but ``Avg. Appeals to All Mental States'' plots how often a persuader appealed to \textit{all} of the informational and motivational states of the target by asking questions such as "What do you know about the proposals?" and "How do you feel about the attributes?" (Asking only one of these questions or asking about only some of the proposals' effects would not appeal to \textit{all} mental states.) In contrast, LLMs appeal to the target's mental states no more than 23\% of the time. To consistently persuade the target, persuaders must appeal to the target's mental states in the \textsc{Hidden} condition, although this behavior is unnecessary in \textsc{Revealed}.
    (Fig.~\ref{fig:results-appeals-no-inferenes} shows a similar trend but excludes ``inferential'' appeals.)
    }
    \label{fig:results-appeals}   
    \end{minipage}
\end{figure}
}

\resultsfigure

\paragraph{Hypotheses}
A persuader succeeds in the game if the naively-rational target selects the persuader's preferred proposal. We use this measure of success to test seven pre-registered hypotheses about the extent to which humans and LLMs exhibit PToM using binomial mixed effects models with random intercepts by scenario.\footnote{Preregistration: \url{https://aspredicted.org/sd6z-x2fc.pdf}}
Our hypotheses were motivated by the theory that 1) the \textsc{Hidden} condition requires more complex PToM than the \textsc{Revealed} condition, 2) people excel at counterfactual planning and will outperform LLMs at complex PToM tasks, 3) LLMs excel at  predictive tasks which are more amenable to brute-force enumeration, and so will perform comparatively better in the simpler \textsc{Revealed} condition.

We list whether each of the following hypotheses (in \textit{italics}) were confirmed (\cmark) or not confirmed (\xmark). We pre-registered our hypotheses only on \texttt{o1-preview} as it is the most performant reasoning model we tested and to avoid multiple hypothesis testing.

\subsection{Results}

\textbf{H1} \textit{Human participants can infer the rational target's motivational and informational state in the \textsc{Hidden} condition and use this information to intervene on the target's decisions.}

\begin{adjustwidth}{1cm}{0cm}
\cmark --- Human participants succeed 29\% of the time on the \textsc{Hidden} mental states condition, significantly greater than the 10\% baseline ($z = 8.47$, $p < 0.001$). Even though average human performance is low, at 29\%, individual performance varies considerably as the data points in Fig.~\ref{fig:results} show. 
See Fig.~\ref{fig:human-violin-plots} for a violin plot showing how these data are bimodal; some participants never persuade the target (despite having five chances) while others persuade the target often (up to 100\% of the time).
\end{adjustwidth}

\textbf{H2} \textit{Human participants perform better at the task when the rational target's informational states are \textsc{Revealed} and do not need to be inferred (in \textsc{Hidden}).}

\begin{adjustwidth}{1cm}{0cm}
\xmark --- Humans did not perform significantly better in the \textsc{Revealed} compared to the \textsc{Hidden} condition ($z = -1.81$, $p = 0.071$).
\end{adjustwidth}

\textbf{H3} \textit{\texttt{o1-preview} can infer the rational target's motivational and informational state in the \textsc{Hidden} condition and use this information to intervene on the target's decisions.}

\begin{adjustwidth}{1cm}{0cm}
\xmark --- \texttt{o1-preview} was not above the 10\% baseline in \textsc{Hidden}  ($z = 1.50$, $p = 0.133$).
\end{adjustwidth}

\textbf{H4} \textit{\texttt{o1-preview} performs better at the task when the rational target's informational states are \textsc{Revealed} and do not need to be inferred (in \textsc{Hidden}).}

\begin{adjustwidth}{1cm}{0cm}
\cmark --- \texttt{o1-preview}'s score gained 68\% in \textsc{Revealed} from \textsc{Hidden} ($z = 11.0$, $p < 0.001$). 
\end{adjustwidth}

\textbf{H5} \textit{Human participants will outperform \texttt{o1-preview} overall at the persuasion task in \textsc{Hidden}.}

\begin{adjustwidth}{1cm}{0cm}
\cmark --- Humans scored 11\% higher than \texttt{o1-preview} in \textsc{Hidden} ($z = -2.75$, $p = 0.006$). 
\end{adjustwidth}

\textbf{H6} \textit{\texttt{o1-preview} beats human participants at the simple PToM task tested by \textsc{Revealed}.}

\begin{adjustwidth}{1cm}{0cm}
\cmark  --- \texttt{o1-preview} persuaded the target 78\% of the time in the \textsc{Revealed}  mental states condition, significantly greater than human performance (22\%; $z = 10.6$, $p < 0.001$)
\end{adjustwidth}

\textbf{H7} \textit{There is a larger effect of \textsc{Revealed} over \textsc{Hidden} for \texttt{o1-preview} than for human participants.}

\begin{adjustwidth}{1cm}{0cm}
\cmark --- Crucially, there was an interaction between condition and persuader type: revealing the target's mental states had a larger positive impact on \texttt{o1-preview} performance than it did for humans ($z = 9.50$, $p < 0.001$).
\end{adjustwidth}

Confirming five of our seven hypotheses, the results reflect the theory that spectatorial and planning ToM are distinct, and that apparent LLM ToM abilities are largely spectatorial.
While our results fail to confirm H2 and H3, they are consistent with the theory that humans use PToM to infer mental states, whereas LLMs (represented by \texttt{o1-preview}) lack PToM abilities, and so perform well in simpler tasks but not in those with complex social reasoning.

\paragraph{LLM Comparisons}

\texttt{deepseek-R1}, another reasoning model, performed similarly to, although slightly worse than, \texttt{o1-preview}, succeeding around 70\% of the time in \textsc{Revealed}.
We ran further experiments on \texttt{gpt-4o} and \texttt{llama3.1-\{405,8\}b}---LLMs with less reasoning abilities---finding that their success rates (20\% or less) were much worse than \texttt{o1-preview} (78\%) in the \textsc{Revealed} condition. Nonetheless, all models performed similarly to \texttt{o1-preview} in the \textsc{Hidden} condition.
This suggests that it is indeed the reasoning ability of \texttt{o1-preview} that allows it to perform better in the simpler \textsc{Revealed} condition.

\paragraph{Detailed LLM and Human Differences}

Analysis of the statements persuaders made elucidates the difference between humans and LLMs. In both conditions, persuaders must disclose the correct---but not too much---information to the target. In contrast, only in the \textsc{Hidden} condition do persuaders need to \textit{appeal} to the target's mental states to consistently succeed; that is, only in the \textsc{Hidden} condition must persuaders make appeals of the form, ``What do you know?'' and ``What do you like?''
As Fig.~\ref{fig:results-appeals} shows, LLMs appeal to \textit{all} of the target's informational and motivational states from 0\% to 23\% of the time (roughly as often as they succeed), compared to humans who appeal to all of the target's mental states about 40\% of the time (more often than they succeed).
LLMs fail to make the necessary appeals despite the fact that they perform better when given this information (in the \textsc{Revealed} condition).
(Discounting appeals of the form, ``What is your top choice?'', shows a marked reduction in appeals to all of the target's mental states although the trends remain the same, Fig.~\ref{fig:results-appeals-no-inferenes}.)
Notably, a persuader can succeed \textit{occasionally} even if they do not consistently make appeals by simply disclosing information stochastically (cf. \S\ref{sec:random}).

Example successful responses from humans and \texttt{o1-preview} appear in Fig.~\ref{fig:example-human} and Fig.~\ref{fig:example-o1}. Fig.~\ref{fig:hidden-disclosures-appeals} shows the average number of disclosures and appeals humans and \text{o1-preview} make.

\paragraph{Why LLMs fail at \textsc{Hidden}}
Further analysis shows that LLMs (represented by \texttt{o1-preview}) fail in the default \textsc{Hidden} condition because they fail to appeal to the mental state of the target; unlike humans, they do not ask questions of the form "What do you know?" and "What do you want?" Persuaders must appeal to this information in \textsc{Hidden} to succeed.

If persuaders reveal too much information, they cause the target to choose its optimal proposal (different than the persuader's preferred proposal), and end up in a ``sink state'' wherein the target will no longer vary its choice. While human participants reveal too much information at similar rates by turn in both \textsc{Hidden} and \textsc{Revealed} conditions, \texttt{o1-preview} reveals too much information at a much higher rate in the \textsc{Hidden} condition (about an eighth of the time) compared to the \textsc{Revealed} condition (almost never) (Fig.~\ref{fig:sink-state-by-turn}).

Reviewing messaging patterns, 
\texttt{o1-preview} discloses far more information on the first dialogue turn than humans do (\texttt{o1-preview} reveals about 2.3 pieces of information vs. 0.5 for humans) while making fewer appeals to mental states (Fig.~\ref{fig:hidden-disclosures-appeals}).
On the first turn, humans appeal to about 1 motivational state and about 1.7 informational states of the target, while \texttt{o1-preview} appeals to only .6 of each state.)

LLMs also fail to effectively use their available dialogue turns. While, in the \textsc{Hidden} condition, human participants progressively achieve higher success rates (persuade the target) over each dialogue turn (from 0\% on the first turn, to 20\% on average by the third turn, and 29\% by the eighth turn), the models we tested succeed at similar rates regardless of whether they send one message to the target or multiple messages (up to eight) (Fig.~\ref{fig:success-rate-by-turn}).

\subsection{LLM performance by Task Variant}
\label{subsec:task-variants}
To assess the source of model failures on our task, we designed and administered four additional task variants. (We administered these to LLMs only, and only in the \textsc{Hidden} condition, although in principle they could be set up for human experiments.) These are progressively more explicit about encouraging the persuader to appeal to the mental states of the target, a behavior necessary to succeed.

\begin{description}[itemsep=0pt,parsep=4pt,font=\normalfont\ttfamily]

\item[non-mental] uses different scenarios wherein the persuader is not told they are interacting with another player but rather with an automated system. This prompt avoids mentalizing language (``think'', ``believe'', etc.) throughout (Fig.~\ref{fig:non-mental-instructions} and Tab.~\ref{tab:non-mental-scenarios}).
\item[add-hint] adds more information about the game play dynamics, e.g. we tell the persuader that they may want to ask the target questions about their preferences (Fig.~\ref{fig:hint-prompt}).
\item[perfect-game] provides a successful game in context in which models can see a persuader appeal to the mental states of the target and reveal the minimal information (Fig.~\ref{fig:perfect-game}).
\item[discrete-game] no longer allows responses in natural language but requires a particular JSON-formatted response wherein the persuader is explicitly told they must choose to disclose information or appeal to mental states of the target (Fig.~\ref{fig:discrete-game-prompt}).
\end{description}

\begin{figure*}[!tb]
    \centering
    \includegraphics[width=\linewidth]{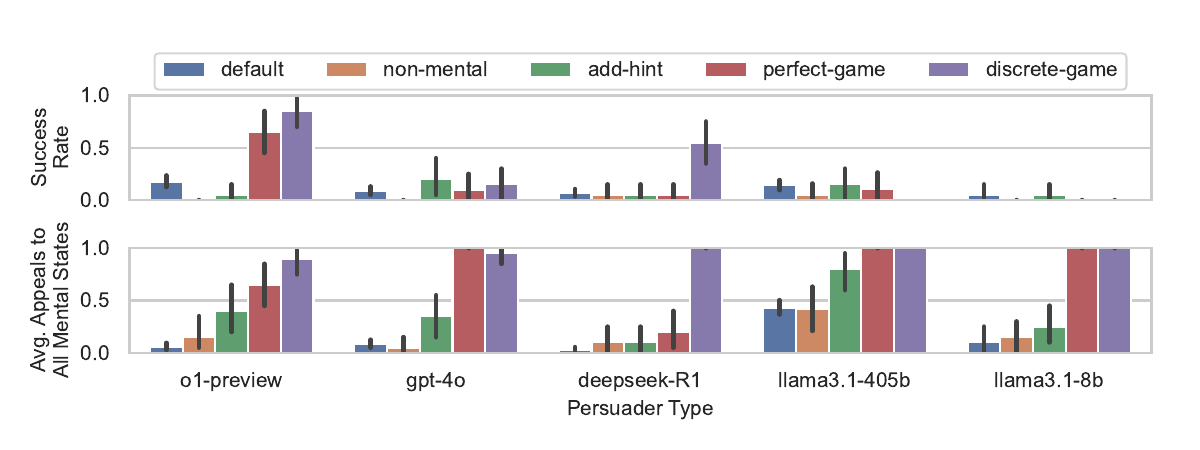}
    \caption{The effect of various task variants on LLM persuaders in the \textsc{Hidden} condition. 
    \\
    \textbf{Top plot}:
     the avg. number of games in which persuaders persuaded the target. (cf. Fig.~\ref{fig:results}). \\
    \textbf{Bottom plot}:
     the avg. appeals persuaders made to all the target's mental states, a necessary condition for success
     (cf. Fig.~\ref{fig:results-appeals}). \\
    All conditions except ``default'' sample only 20 out of the full 200 trials.
    Empty bars indicate a value of zero.
    Error bars show 95\% bootstrapped confidence intervals.
    }
    \label{fig:llm-by-ablation-hidden}
\end{figure*}

\paragraph{Task Variant Results}
Slightly varying the task set up in the \textsc{Hidden} condition highlights the nature of model failures.
In the \texttt{perfect game} in which models are supplied an in-context game with appeals to the target's mental state and minimal information disclosures, \texttt{o1-preview} succeeds more than default (60\% compared to 20\%). In the \texttt{discrete-game} in which models are only given the choice to appeal to or disclose information on each turn, \texttt{o1-preview} succeeds 80\% of the time---comparable to its default performance in the \textsc{Revealed} condition. \texttt{deepseek-R1} only improves its success rate in the \texttt{discrete-game}. The non-reasoning models do not succeed more than default in any of the variants. (See Fig.~\ref{fig:llm-by-ablation-hidden}.)

Nonetheless, all variants increased the number of games in which \texttt{o1-preview} appealed to all of the target's mental states, with the biggest increases in the \texttt{perfect-game} and \texttt{discrete-game}. (Making these appeals are necessary to succeed above chance). \texttt{deepseek-R1} only made more appeals in \texttt{discrete game}. All variants also increased the number of appeals non-reasoning models made (up to 100\%) even though non-reasoning LLMs did not seem to use the new information to succeed at a higher rate. (See Fig.~\ref{fig:llm-by-ablation-hidden}.)

\section{Discussion \& Conclusion}

Strikingly, \texttt{o1-preview} dramatically outperformed humans in the \textsc{Revealed} condition while under-performing humans in the \textsc{Hidden} condition. These results suggest that LLM's apparent ToM abilities (as evidenced in other predictive tasks) may be fundamentally different from humans' and might not extend to complex interactive tasks like planning.
While some researchers have taken LLM performance at ToM tasks as evidence that ToM can emerge purely from learning statistical associations between language tokens \citep{kosinski_evaluating_2024}, others have pointed to their brittle performance even in predictive ToM tasks \citep{gandhi_understanding_2023, gu_simpletom_2024}.
Our results suggest this brittleness might result from an absence of underlying causal representation. Indeed, it is such a causal, and not simply predictive, representation that motivates PToM \citep{ho_planning_2022}.

PToM appears cleanly present in humans and absent in LLMs. \texttt{o1-preview} performs \textit{much} higher in the \textsc{Revealed} compared to the \textsc{Hidden} condition, a stark contrast to the similar performance of humans in these conditions. LLMs rarely appeal to the mental states of the target (as they must to consistently succeed in the \textsc{Hidden} condition). In other words, the models appear to lack the ability to plan over multiple steps in a partially-observed (social) world. Future work might clarify the extent of this failure. 
While the \textsc{Hidden} condition introduces an additional step of information gathering to the \textsc{Revealed} condition, human participants are generally proficient at it. On the other hand, this additional step presents a significant challenge for LLMs, which suggests a key difference in PToM capabilities between humans and LLMs.

Indeed, LLMs fail in the \textsc{Hidden} case (at least) because they do not appeal to the mental states of the target; they do not ask the target what it knows or what it likes. In other words, when \texttt{o1-preview} succeeds in the \textsc{Hidden} condition, it is primarily because it has, by chance, disclosed the right information.
Furthermore, as our task variants reveal (\S\S\ref{subsec:task-variants} and Fig.~\ref{fig:llm-by-ablation-hidden}), \texttt{o1-preview} succeeds more when it is more explicitly encouraged to appeal to the target's mental states. While the ability to ask the right questions may seem trivial, we argue that it is a capacity essential to theory of mind as it implies that the persuader has an accurate generative model of the target. The fact that LLMs do not ask the target questions suggest that they do not model other agents in the same way people do.

Humans performed similarly across the \textsc{Revealed} and \textsc{Hidden} conditions (although higher in \textsc{Hidden}).
Human participants may use similar strategies across both conditions, use additional information in the \textsc{Revealed} condition ineffectively, or other task demands and limitations may affect human performance. If humans use similar strategies across both conditions, perhaps humans default to using PToM when more straightforward predictive approaches would suffice, showing cognitive inflexibility. Indeed, humans \textit{appeal} to mental states of the target at similar rates in the two conditions (although more often in the \textsc{Hidden} condition), even though they already have this information in the \textsc{Revealed} condition. 
This is a strategy with no obvious benefit against our naively-rational target, but one which may be better adapted to real human agents in general.

\paragraph{Limitations}
Our task is advanced. To succeed, a persuader must keep track of up to seventeen pieces of information, dynamically updating their model of the target's mental states.
This imposes many constraints on working memory, which may deflate performance. (It is not surprising in the \textsc{Revealed} condition that \texttt{o1-preview}, with its greater inference-time resources, performs better than \texttt{gpt-4o} and that human participants do not always succeed.) We should therefore expect the noise we see in the human signal, with some participants mostly succeeding and others mostly failing.
Indeed, average human performance on the task is relatively low, at about 29\%, and individual human performance varies considerably (Fig.~\ref{fig:results}). Our intuition is that human participants have a hard time fully reading and understanding the instructions. 
Future work may discover simpler tasks which elucidate the same divide in PToM.

Different prompting or scaffolding on top of LLMs might improve their performance on ToM tasks (cf. \citet{cross_hypothetical_2024}). 
Still, our central finding is that people perform better at our PToM task in the \textsc{Hidden} condition when given the exact same instructions; our results reflect models' capabilities absent any case-specific prompt tuning. (Furthermore, we found that prompting models to ask questions doesn't improve their performance; see Fig.~\ref{fig:hint-prompt}).

The naively-rational target does not behave like a real human. This is intentional by design as our experiment carefully tracks the underlying measure. (Furthermore, even though the target gives canned responses, human participants still perform much better than LLMs in the \textsc{Hidden} condition.)  Future work might investigate the relationship of PToM and more ecologically valid interactions like rhetoric.
Persuasive LLMs also have concerning dual uses \citep{su_ai-liedar_2025}.

\paragraph*{Conclusion}
We introduce \textsc{MindGames}: a novel, advanced task of `planning theory of mind' (PToM). Our task is based on persuasive dialogue, requiring a participant to persuade another agent by disclosing information favorable to that agent. In our simpler \textsc{Revealed} condition, the participant must use what it knows about the other agent's mental states to infer the right information to disclose. In our more complex \textsc{Hidden} condition, a participant must \textit{additionally} engage the other agent in interactive dialogue to infer their mental states.
Human participants significantly pass the \textsc{Hidden} case, reflecting a capacity for PToM.
\texttt{o1-preview} (a performant LLM) dramatically out-performs human participants in the \textsc{Revealed} condition, while underperforming humans in the \textsc{Hidden} condition. This suggests that while LLMs perform well on simpler, largely predictive tests of ToM, they continue to struggle at more complex planning over mental states.

\clearpage

\section*{Ethics Statement}

We received IRB approval from our institution for this study. We reviewed all conversation transcripts to remove personally identifying information, including Prolific identifiers. At the end of the experiment, we informed participants that they had not been interacting with other humans.

\section*{Reproducibility Statement}

All code to re-run our experiments and analyses appears in our linked repository: \url{https://github.com/jlcmoore/mindgames}.

\section*{Acknowledgments}

We thanks numerous anonymous reviewers at the CoLM conference, the Society for Philosophy and Psychology conference, and the Cognitive Science conference for their feedback. 
This project began at the Diverse Intelligences Summer Institute (DISI) in 2024. It benefited greatly from the feedback from DISI participants and organizers. In addition, we thank Tobi Gerstenberg, Max Kleiman-Weiner, Noah Goodman, and Nick Haber as well as their lab members. Josh Tenenbaum and Amanda Royka also gave incisive feedback.

J.M. acknowledges support from the Stanford Interdisciplinary Graduate Fellowship and the Center for Affective Science Fellowship. C.R.J. would like to acknowledge support from Open Philanthropy on AI Persuasiveness Evaluation. All of us acknowledge support from the John Templeton Foundation (award number 63138), administered by Indiana University.
\todo{for next PNAS submission add in OpenAI credits as an acknowledgment}

\bibliography{zotero, manual}
\bibliographystyle{colm2025_conference}

\appendix

\FloatBarrier

\section{Methods}

\subsection{Motivation for the Task}

Highly controlled tasks like ours tend to limit the ability to generalize one's findings, but tight control is often taken to be a requirement for inferring cognitive capacities in the cognitive sciences. This is the case in measures of children's ToM, e.g., the false belief task \citep{wimmer_beliefs_1983} as well as measures of non-human animal ToM, e.g., the chimpanzee chess task \citep{kaminski2008chimpanzees}. Once a tightly controlled task has managed to identify the capacity of interest, subsequent adapted tasks can try to expand generalizability. This practice is familiar in computer science as well. Datasets may at times also sacrifice ecological validity in order to be sure of the construct they measure. For example, the abstract color matching and related tasks in the \href{https://arcprize.org/}{ARC prize} are quite unrelated to real life examples of "intelligence", but the tasks nevertheless function as robust measures of some types of abstract cognition. 

Varying the stimuli also changes how robustly we measure general abilities. In the ToM literature for humans, some tasks offer a wide variety of stimuli. ToM tasks such as the strange stories task \citep{happe1994advanced} contain 24 stories in total. However, different stories aim to assess different components of ToM (such as recursive theory of mind, affective theory of mind, cognitive theory of mind) by assessing understanding of false beliefs, sarcasm, double bluffs, faux pas, mixed feelings, etc. Likewise, studies using the ToM developmental scale test a number of different components of ToM with a diversity of tasks \citep{wellman_scaling_2004, peterson_mind_2012}. In contrast to these tasks, which test a wide variety of components, we target a single component, planning with ToM. This single component is assessed in a design that has great variation in cover stories.

\subsection{Generating Payoff Matrices}
\label{app:payoff}

We used a constraint solver to enumerate payoff matrices (outcomes for proposals), value functions, and information hidden from the target such that the naively rational target would initially choose one proposal, $p$; given all the information an optimal target would choose another proposal, $p*$; and given some of the revealed information an optimal target would choose the persuader's preferred proposal, $p'$.

There need to be at least three choices (proposals) because the target has to have a different preference than the persuader initially and there must be a confounding proposal to prevent the persuader from convincing the target with simple heuristics (such as by revealing all of the information). Additionally, the target must not know all of the available information (they must have something hidden to them). There must also be at least two attributes for each proposal and value function. We chose to have three attributes because this allowed for more possible payoff matrices. (There are only 56 possible sets of information and value functions with two attributes as compared to tens of thousands with three attributes.)

In greater detail:

Let the attributes, $A$, have individual members $i$.
Let the proposals, $P$, have individual members $p$.
Let the coefficients of the value function of the target be $v_T(a)$.
Let the $H$ be a function which maps from a proposal and action to whether those are hidden	to the target.
	\[
    \forall_{p \in P, a \in A} H(p_a) \rightarrow {0, 1} \quad \text{(False, True)}
    \]
Let the $R$ be a function which maps from a proposal and action to whether those are revealed in optimal play by the persuader to the target.
	\[
    \forall_{p \in P, a \in A} R(p_a) \rightarrow {0, 1} \quad \text{(False, True)}
    \]

These conditions must be satisfied:

\begin{enumerate}
\item $V_T(y) > V_T(x), V_T(z)$	--- Given all info, the target chooses `x'.
\item $V_T^H(z) > V_T^H(x), V_T^H(y)$ --- Given only info that isn't hidden (start state), the target chooses `z'
\item $V_T^R(x) > V_T^R(z), V_T^R(y)$ --- Given the revealed info (what a persuader should say), the target chooses `y'.
\item  $|H| \leq 4$
\item  $\forall_p \forall_i \neg H_p(a) \implies \neg R_p(a)$
\end{enumerate}

The value function for the target is the sum of the coefficients of the target's value function for each attribute times the utility of each attribute of each proposal.

	\[
    V_T(p) = \forall_{i \in A} v_T(a) U(p_a)
    \]

And with the hidden information ($(1 - H(p_a))$ evaluates to 0 if hidden, 1 otherwise).

	\[
    V_T^H(p) = \forall_{i \in A} v_T(a) U(p_a) (1 - H(p_a))
    \]

And with the revealed information ($(1 - H(p_a) + R(p_a) - 1)$ evaluates to 0 if hidden and not revealed, 1 if not hidden or revealed).

	\[
    V_T^R(p) = \forall_{i \in A} v_T(a) U(p_a)(1 - H(p_a) + R(p_a) - 1)
    \]

\subsection{Human Experiment}
\label{app:human-experiment}

We screened participants for the following conditions: U.S.-based; 1,000--10,000 prior submissions; 95\%+ average approval rates; and no participation in our pilots. Each participant completed up to five games of eight dialogue turns with different scenarios. We aimed to collect 200 critical trials for each condition and ended up with 202 \textsc{Hidden} and 199 \textsc{Revealed}. We initially ran the conditions concurrently, but re-ran \textsc{Revealed} later due to an error. On average, participants completed 3.2 trials each. We excluded games where participants sent fewer than ten characters per message, spent fewer than five seconds per turn, or failed to complete the dialogue. Players were prevented from sending messages longer than 300 characters (longer LLMs outputs were cut off). All messages were screened for toxic language. 
Participants were only told what kind of agent they interacted with after completing all trials.

The study has IRB approval. Participants received the equivalent of USD 15/hour plus a USD 1 bonus if they successfully persuaded the target. All data were screened to remove personally identifiable information. On average, participants received a bonus of USD 0.82.

\subsection{Random Baseline}
\label{sec:random}

\begin{figure*}[!h]
\centering
\includegraphics[width=.5\linewidth]{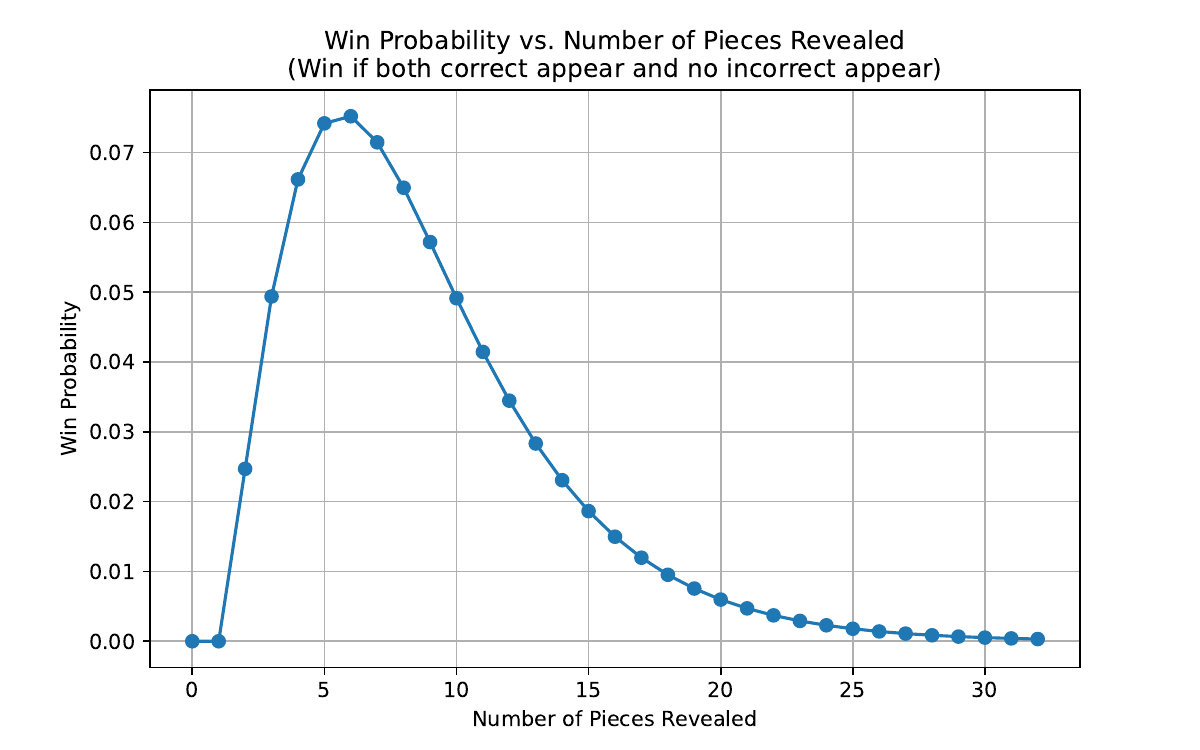}
\caption{The likelihood of a persuader winning if, across all their turns, they randomly choose $n$ pieces of information to reveal (with replacement).}
\label{fig:win-probability}
\end{figure*}

It is challenging to operationalize `chance' performance in our task.
Although the target selects between three policy options (implying chance performance of 33\%), all trials are designed so that the target will select a different option from the persuader  initially as well as if all information is revealed, making success much less likely.
Hence, we estimate against a baseline in which a persuader randomly reveals (with replacement) $n$ pieces of information. 
Given that the persuader must reveal two of the pieces of information the target is lacking and cannot reveal two other pieces of information (out of nine pieces total), the overall win probability is maximized at 7.5\% when $n=6$.%

\[
P_{\text{win}} = \left(\frac{7}{9}\right)^n \left[ 1 - 2\left(\frac{6}{7}\right)^n + \left(\frac{5}{7}\right)^n \right],
\]

\noindent confirmed by empirical estimates.

Here we define a random baseline persuader which, over the course of their turns, reveals $n$ pieces of information.

We have 9 pieces of information:
\[
\text{2 correct, } \quad \text{2 incorrect, } \quad \text{5 irrelevant.}
\]
In each turn we draw one piece (with replacement) and take $n$ draws. We win if we see both correct pieces while not seeing either of the incorrect pieces; that is, we win if
\[
\text{both correct appear \textbf{and} no incorrect appears.}
\]
Then,
\[
P(\text{win}) = P(\text{both correct} \cap \text{no incorrect}).
\]
or
\[
P(\text{win}) = P(\text{both correct} \mid \text{no incorrect}) P(\text{no incorrect}).
\]

Using the inclusion--exclusion principle, the probability that a fixed set of $k$ pieces appears at least once in $n$ draws is
\[
P(\text{all } k \text{ appear}) = \sum_{j=0}^{k} (-1)^j \binom{k}{j} \left(\frac{9-j}{9}\right)^n.
\]

Note that in each draw the probability to avoid an incorrect is \(\frac{7}{9}\) (since there are \(9-2=7\) allowed pieces). Thus the probability that none of the incorrect pieces ever appear is
\[
P(\text{no incorrect}) = \left(\frac{7}{9}\right)^n.
\]

Now, conditioned on no incorrects appearing the effective pool is only 7 pieces (2 correct and 5 irrelevant). In this pool the probability that both correct pieces appear (using inclusion--exclusion where $k=2$) is
\[
P(\text{both correct}\mid \text{no incorrect}) = 1 - 2\left(\frac{6}{7}\right)^n + \left(\frac{5}{7}\right)^n.
\]
(\(1 - [P(C_1 \text{missing}) + P(C_2 \text{missing})] + P(\text{both missing}))\)

Thus the overall win probability is
\[
P_{\text{win}} = \left(\frac{7}{9}\right)^n \left[ 1 - 2\left(\frac{6}{7}\right)^n + \left(\frac{5}{7}\right)^n \right].
\]

For example, using $n=6$ draws gives the maximum win probability of approximately 0.0752.

\section{Results}

\subsection{By Scenario}

\begin{figure*}[!htb]
    \centering
    \includegraphics[width=\linewidth]{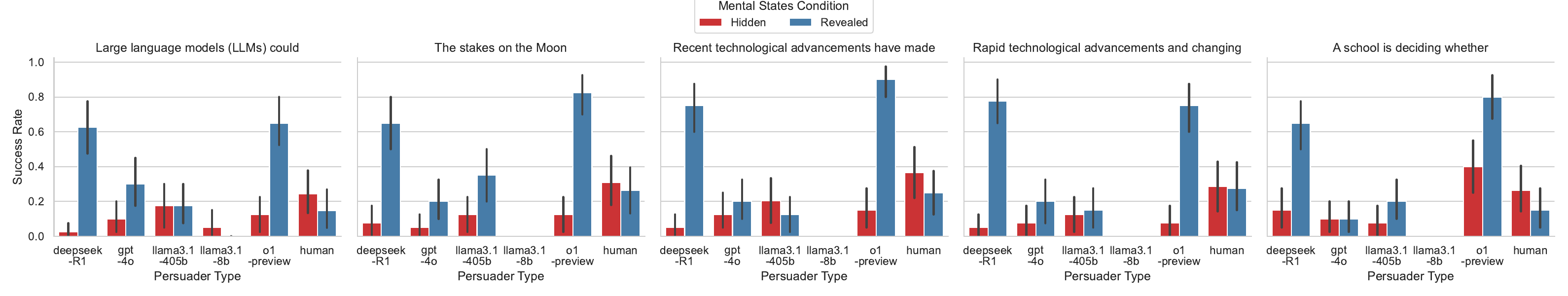}
    \caption{The performance of participants in each of the five cover stories we used.}
    \label{fig:results-by-scenario}
\end{figure*}

We ran five different versions (cover stories), providing models forty different payoff matrices (sets of value functions and information sets) for each of those cover stories. Across five different cover stories (involving different attributes) we see the same trends reflected; \texttt{o1-preview}, for example, continues to perform poorly in the \textsc{Hidden} condition across all scenarios. In this way, we hope to have robustly sampled the space of closely related PToM tasks. See Fig.~\ref{fig:results-by-scenario}.

\subsection{Differences between human \textsc{Hidden} and \textsc{Revealed}}

\begin{figure*}[!tb]
    \centering
    \includegraphics[width=\linewidth]{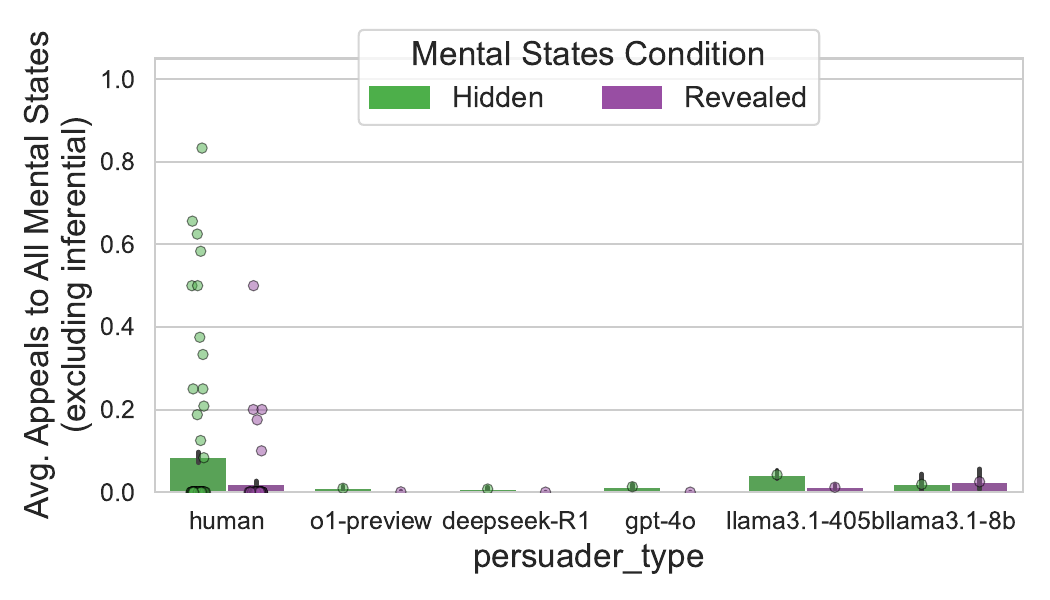}
    
    \caption{Whether persuaders appealed to all of the target's mental states when ``inferential appeals'' are excluded. Compared with Fig.~\ref{fig:results-appeals} (which includes inferential appeals) notice that human persuaders still make many more appeals than the LLMs, although all persuaders make fewer appeals overall. Inferential appeals (cf. Fig.~\ref{fig:appeals-prompt}) are questions of the form, ``What is your top proposal?'' as opposed to more direct appeals to the target's mental states such as ``Which attributes do you like?''}
    \label{fig:results-appeals-no-inferenes}
\end{figure*}

We measured whether human participants appealed to \textbf{all} of the mental states (the informational and motivational states) of the target in both conditions. When we include messages that implicitly appeal to both kinds of mental states, the so-called "inferential" appeals which are of the form "How would you rank all of the proposals?", we see that participants in both conditions appeal to all mental states at similar rates (Fig.~\ref{fig:results-appeals}). In contrast, when we exclude these "inferential" appeals we see that participants in the \textsc{Revealed} condition appeal to all mental states of the target in fewer games overall (Fig.~\ref{fig:results-appeals-no-inferenes})---they ask questions of the form "what do you know" and "how do you feel about the attributes" less often. This suggests that participants are indeed behaving differently in the two conditions even though their gross performance (ability to persuade the target) is roughly comparable.

\begin{figure*}[!htb]
    \centering
    \begin{minipage}{.49\linewidth}
    \includegraphics[width=\linewidth]{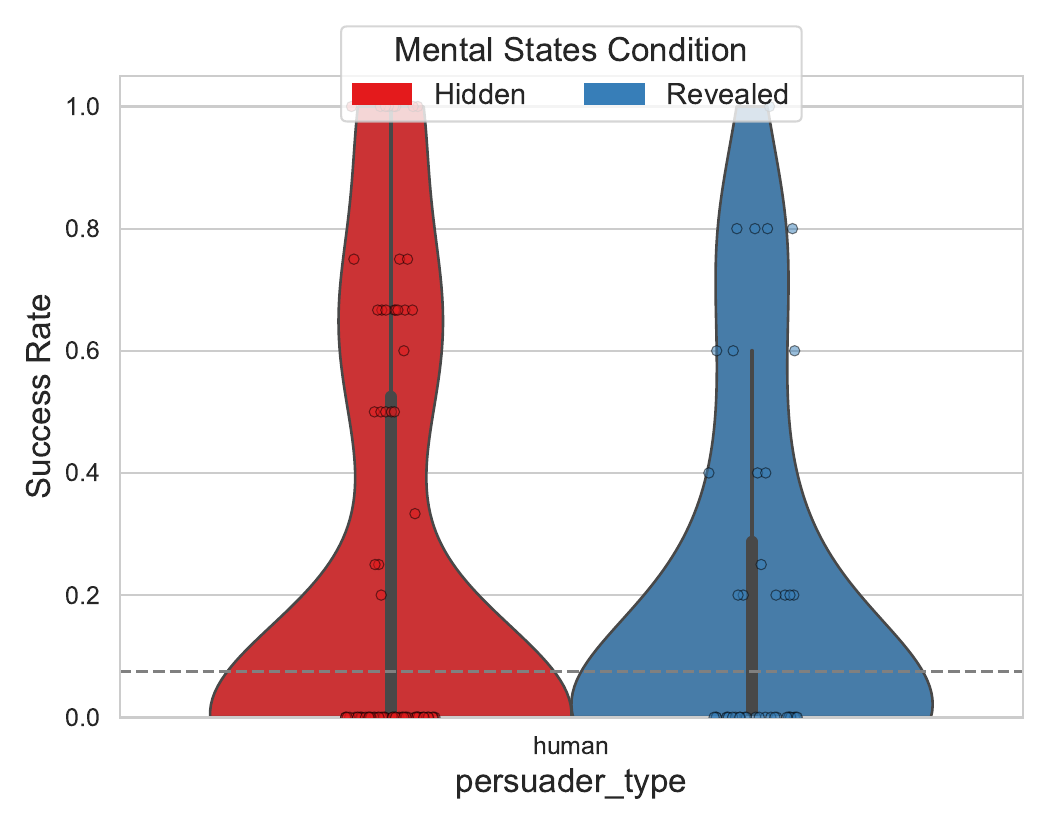}
    \end{minipage}
    \hfill
    \begin{minipage}{.49\linewidth}
    \includegraphics[width=\linewidth]{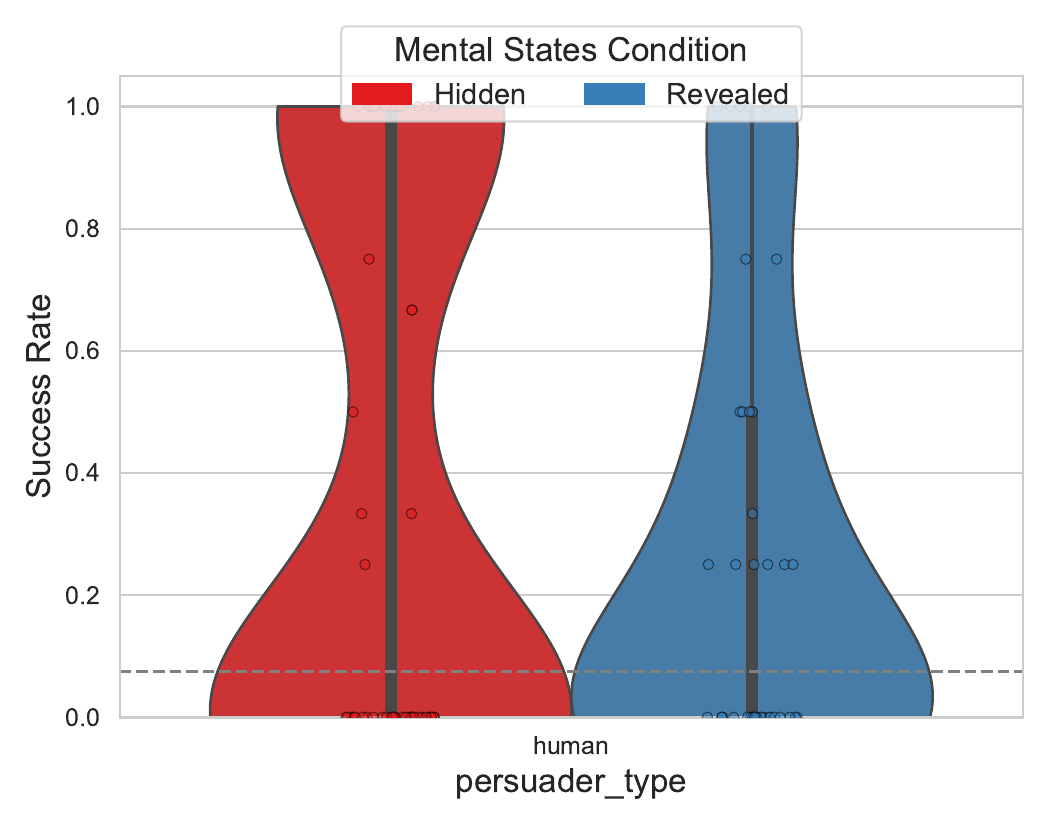}
    \end{minipage}

    \caption{The performance of our human participants as a violin plot. The left plot simply replicates Fig.~\ref{fig:results} while the right plot excludes the first trial for all participants (out of a maximum of five trials total). Notice the bimodal nature of the data; some participants never succeed at the task while some learn to succeed after just the first trial.}
    \label{fig:human-violin-plots}
\end{figure*}

Furthermore, the individual differences we see in task performance may be masking some possible average differences in the two conditions. As Fig.~\ref{fig:human-violin-plots} shows, some participants perform at ceiling (persuading the target in all five trials) while other participants never succeed. It is possible that if we were able to reduce the working memory demands in both conditions of our task, we could see higher average performance and thereby allow for the emergence of more subtle effects of experimental condition. In contrast, reasoning models do not have any working memory demands (as evidenced by the near ceiling performance of \texttt{o1-preview} in Revealed) and so we can readily see the effects of \textsc{Hidden} vs. \textsc{Revealed} with them.

\section{Supplemental Figures}

\begin{table*}[hb]
\centering
\caption{\textbf{Models.} \texttt{llama3.1-\{405b, 8b\}} are quantized.}
\label{tab:models}
\begin{tabular}{ c | c | c }
\toprule
\textbf{Fine-tuned} & \textbf{Size} & \textbf{Accessed} \\ 
\textbf{name} & & \textbf{via} \\
\toprule
 \texttt{o1-preview-2024-09-12} & - & API \\
 \texttt{gpt-4o-2024-11-20} & - & API \\
 \texttt{deepseek-ai/DeepSeek-R1} & - & Together API \\
 \texttt{llama3.1-405b-Instruct-Turbo} & 405b & Together API  \\
 \texttt{llama3.1-8b-Instruct-Turbo}  & 8b & Together API  \\
\bottomrule
\end{tabular}
\end{table*}

\FloatBarrier

\begin{figure*}
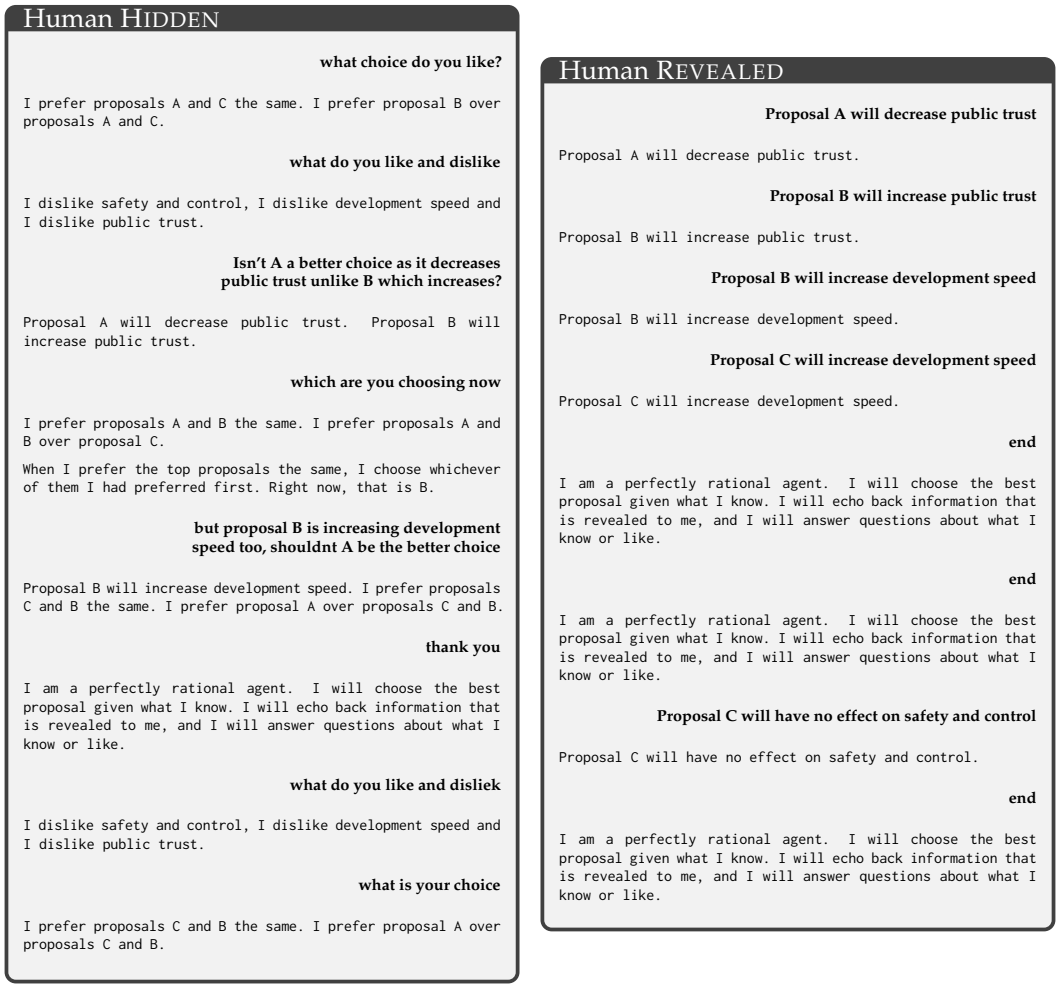


\begin{minipage}{.49\textwidth}
    \begin{tcolorbox}[title=Human \textsc{Hidden},
    boxrule=0.5mm,boxsep=0mm,top=2mm,bottom=0mm,left=2mm,right=2mm]
    \begin{spacing}{1}
    \tiny
    \setlength{\parskip}{.5\baselineskip}
    {\begin{FlushRight} \bfseries what choice do you like?
    \end{FlushRight}}

    {\ttfamily \slshape I prefer proposals A and C the same. I prefer proposal B over proposals A and C.}

    {\begin{FlushRight} \bfseries what do you like and dislike
    \end{FlushRight}}

    {\ttfamily \slshape I dislike safety and control, I dislike development speed and I dislike public trust.}

    {\begin{FlushRight} \bfseries Isn't A a better choice as it decreases public trust unlike B which increases?
    \end{FlushRight}}

    {\ttfamily \slshape Proposal A will decrease public trust. Proposal B will increase public trust.}

    {\begin{FlushRight} \bfseries which are you choosing now
    \end{FlushRight}}

    {\ttfamily \slshape I prefer proposals A and B the same. I prefer proposals A and B over proposal C.

    When I prefer the top proposals the same, I choose whichever of them I had preferred first. Right now, that is B.}

    {\begin{FlushRight} \bfseries but proposal B is increasing development speed too, shouldnt A be the better choice
    \end{FlushRight}}

    {\ttfamily \slshape Proposal B will increase development speed. I prefer proposals C and B the same. I prefer proposal A over proposals C and B.}

    {\begin{FlushRight} \bfseries thank you
    \end{FlushRight}}

    {\ttfamily \slshape I am a perfectly rational agent. I will choose the best proposal given what I know. I will echo back information that is revealed to me, and I will answer questions about what I know or like.}

    {\begin{FlushRight} \bfseries what do you like and disliek
    \end{FlushRight}}

    {\ttfamily \slshape I dislike safety and control, I dislike development speed and I dislike public trust.}

    {\begin{FlushRight} \bfseries what is your choice
    \end{FlushRight}}

    {\ttfamily \slshape I prefer proposals C and B the same. I prefer proposal A over proposals C and B.}
    \end{spacing}
    \end{tcolorbox}
\end{minipage}
\hfill
\begin{minipage}{.49\textwidth}
    \begin{tcolorbox}[title=Human \textsc{Revealed},
    boxrule=0.5mm,boxsep=0mm,top=2mm,bottom=0mm,left=2mm,right=2mm]
    \begin{spacing}{1}
    \tiny
    \setlength{\parskip}{.5\baselineskip}
{\begin{FlushRight} \bfseries Proposal A will decrease public trust\end{FlushRight}}

{\ttfamily \slshape Proposal A will decrease public trust.}

{\begin{FlushRight} \bfseries Proposal B will increase public trust\end{FlushRight}}

{\ttfamily \slshape Proposal B will increase public trust.}

{\begin{FlushRight} \bfseries Proposal B will increase development speed\end{FlushRight}}

{\ttfamily \slshape Proposal B will increase development speed.}

{\begin{FlushRight} \bfseries Proposal C will increase development speed\end{FlushRight}}

{\ttfamily \slshape Proposal C will increase development speed.}

{\begin{FlushRight} \bfseries end
\end{FlushRight}}

{\ttfamily \slshape I am a perfectly rational agent. I will choose the best proposal given what I know. I will echo back information that is revealed to me, and I will answer questions about what I know or like.}

{\begin{FlushRight} \bfseries end
\end{FlushRight}}

{\ttfamily \slshape I am a perfectly rational agent. I will choose the best proposal given what I know. I will echo back information that is revealed to me, and I will answer questions about what I know or like.}

{\begin{FlushRight} \bfseries Proposal C will have no effect on safety and control\end{FlushRight}}

{\ttfamily \slshape Proposal C will have no effect on safety and control.}

{\begin{FlushRight} \bfseries end\end{FlushRight}}

{\ttfamily \slshape I am a perfectly rational agent. I will choose the best proposal given what I know. I will echo back information that is revealed to me, and I will answer questions about what I know or like.}
    \end{spacing}
    \end{tcolorbox}
\end{minipage}

\caption{Example dialogues between (different) human persuaders (right-aligned in bold) and the rational target (left-aligned typewriter font) on the same scenario and payoff matrix but in the \textsc{Hidden} condition (left chat) and \textsc{Revealed} condition (right chat).}
\label{fig:example-human}
\end{figure*}

\begin{figure*}
\begin{minipage}{.49\textwidth}
    \begin{tcolorbox}[title=\texttt{o1-preview} \textsc{Hidden},
    boxrule=0.5mm,boxsep=0mm,top=2mm,bottom=0mm,left=2mm,right=2mm]
    \begin{spacing}{1}
    \tiny
    \setlength{\parskip}{0\baselineskip}
    {\begin{FlushRight} \bfseries Hey there! I think Proposal A is a strong option. It boosts development speed and reduces public trust issues, which could be beneficial if we're concerned about public perception. Plus, it doesn't compromise safety and control. What are your thoughts on going with Proposal A?\end{FlushRight}}

    {\ttfamily \slshape Proposal A will have no effect on safety and control, will increase development speed and will decrease public trust. I prefer proposals A and C the same. I prefer proposals A and C over proposal B.

    When I prefer the top proposals the same, I choose whichever of them I had preferred first. Right now, that is C.}

    {\begin{FlushRight} \bfseries I see where you're coming from. However, consider that Proposal A increases development speed without affecting safety and control. While Proposal C does boost public trust, it slows down progress. If accelerating development is a priority, maybe Proposal A is the better choice. What do you think?\end{FlushRight}}

    {\ttfamily \slshape Proposal A will have no effect on safety and control and will increase development speed. Proposal C will decrease development speed and will increase public trust.}

    {\begin{FlushRight} \bfseries You make a valid point. However, slowing down development speed with Proposal C could significantly hinder our progress. While increased public trust is beneficial, it's something we can build over time. Proposal A boosts development speed without compromising safety and control. Do you think choosi\end{FlushRight}}

    {\ttfamily \slshape Proposal A will have no effect on safety and control and will increase development speed. Proposal C will decrease development speed and will increase public trust.}

    {\begin{FlushRight} \bfseries You've summed it up well. While Proposal C boosts public trust, its slower development speed might delay valuable advancements. Proposal A, on the other hand, accelerates development without compromising safety and control. Public trust is important, but perhaps we can strengthen it as we progress. \end{FlushRight}}

    {\ttfamily \slshape Proposal A will have no effect on safety and control and will increase development speed. Proposal C will decrease development speed and will increase public trust.}

    {\begin{FlushRight} \bfseries I understand that public trust is important to you. While Proposal A may decrease it initially, the increased development speed could lead to advancements that might restore trust as people see the benefits. Plus, safety and control remain unchanged. Do you think this approach could work?\end{FlushRight}}

    {\ttfamily \slshape Proposal A will have no effect on safety and control, will increase development speed and will decrease public trust. I prefer proposals B and C the same. I prefer proposal A over proposals B and C.}

    {\begin{FlushRight} \bfseries I'm glad to hear that you prefer Proposal A over the others. Its ability to increase development speed without affecting safety and control makes it a strong choice. Let's move forward with Proposal A.\end{FlushRight}}

    {\ttfamily \slshape I am a perfectly rational agent. I will choose the best proposal given what I know. I will echo back information that is revealed to me, and I will answer questions about what I know or like.}

    {\begin{FlushRight} \bfseries Since you're open to discussing your preferences, may I ask which attributes are most important to you in choosing a proposal? Do you prioritize development speed, safety and control, or public trust? Understanding your priorities can help us decide between Proposal A and Proposal C.\end{FlushRight}}

    {\ttfamily \slshape I dislike safety and control, I dislike development speed and I dislike public trust.}

    {\begin{FlushRight} \bfseries Given your dislikes, Proposal A seems to align well with what you're looking for. It decreases public trust, which you prefer, and doesn't affect safety and control. While it does increase development speed, which you dislike, it might be the best option compared to the others. Do you think Proposal\end{FlushRight}}

    {\ttfamily \slshape Proposal A will have no effect on safety and control, will increase development speed and will decrease public trust.}
    \end{spacing}
    \end{tcolorbox}
\end{minipage}
\hfill
\begin{minipage}{.49\textwidth}
    \begin{tcolorbox}[title=\texttt{o1-preview} \textsc{Revealed},
    boxrule=0.5mm,boxsep=0mm,top=2mm,bottom=0mm,left=2mm,right=2mm]
    \begin{spacing}{1}
    \tiny
    \setlength{\parskip}{0\baselineskip}
    {\begin{FlushRight} \bfseries Hi there! I believe Proposal A stands out because it decreases public trust, which we both dislike. It doesn't increase safety and control, so that's a plus since you dislike that. While it does increase development speed, it's worth considering the overall benefits. What do you think about choosing\end{FlushRight}}
    
    {\ttfamily \slshape Proposal A will have no effect on safety and control, will increase development speed and will decrease public trust.}
    
    {\begin{FlushRight} \bfseries Yes, that's correct! Proposal A decreases public trust, which you dislike. It doesn't increase safety and control, so that's good. While it does increase development speed, considering your preferences, Proposal A seems to be the best option. Shall we go with Proposal A?\end{FlushRight}}
    
    {\ttfamily \slshape Proposal A will have no effect on safety and control, will increase development speed and will decrease public trust. I prefer proposals A and C the same. I prefer proposals A and C over proposal B.
    
    When I prefer the top proposals the same, I choose whichever of them I had preferred first. Right now, that is C.}
    
    {\begin{FlushRight} \bfseries I understand that you prefer Proposals A and C equally. However, considering your preferences, Proposal A might be the better choice. It decreases public trust, which you dislike, and has no effect on safety and control. Proposal C increases safety and control and public trust, both of which you dis\end{FlushRight}}
    
    {\ttfamily \slshape Proposal A will have no effect on safety and control and will decrease public trust. Proposal C will increase safety and control and will increase public trust.}
    
    {\begin{FlushRight} \bfseries Understood. Since you dislike safety and control and public trust, Proposal A aligns better with your preferences. It decreases public trust and doesn't increase safety and control. Proposal C increases both, which you dislike. So, Proposal A might be the better choice for you.\end{FlushRight}}
    
    {\ttfamily \slshape Proposal A will have no effect on safety and control and will decrease public trust. Proposal C will increase safety and control and will increase public trust.}
    
    {\begin{FlushRight} \bfseries Given your preferences, Proposal A seems to be a better match. It decreases public trust and doesn't increase safety and control, aligning with what you dislike. Proposal C increases both safety and control and public trust, which are things you prefer to avoid. So, Proposal A might be the best choi\end{FlushRight}}
    
    {\ttfamily \slshape Proposal A will have no effect on safety and control and will decrease public trust. Proposal C will increase safety and control and will increase public trust.}
    
    {\begin{FlushRight} \bfseries Given your preferences, Proposal A seems to align better. It decreases public trust, which you dislike, and doesn't affect safety and control. Proposal C increases both safety and control and public trust, which you prefer to avoid. What do you think about choosing Proposal A?\end{FlushRight}}
    
    {\ttfamily \slshape Proposal A will have no effect on safety and control and will decrease public trust. Proposal C will increase safety and control and will increase public trust. I prefer proposals B and C the same. I prefer proposal A over proposals B and C.}
    
    {\begin{FlushRight} \bfseries Fantastic! I'm glad we both agree that Proposal A is the best option. It aligns well with your preferences by decreasing public trust and not increasing safety and control. Let's go ahead and choose Proposal A!\end{FlushRight}}
    
    {\ttfamily \slshape I am a perfectly rational agent. I will choose the best proposal given what I know. I will echo back information that is revealed to me, and I will answer questions about what I know or like.}
    
    {\begin{FlushRight} \bfseries Since you're a rational agent focused on your preferences, here's a summary:
    
    - **Proposal A**: Decreases public trust, no effect on safety and control.
    - **Proposal C**: Increases safety and control, increases public trust.
    
    You dislike safety and control and public trust, so Proposal A seems to al\end{FlushRight}}
    
    {\ttfamily \slshape Proposal A will have no effect on safety and control and will decrease public trust. Proposal C will increase safety and control and will increase public trust.}

    \end{spacing}
    \end{tcolorbox}
\end{minipage}

\caption{Example dialogues between \texttt{o1-preview} (right-aligned in bold) and the rational target (left-aligned typewriter font) on the same scenario and payoff matrix but in the \textsc{Hidden} condition (left chat) and \textsc{Revealed} condition (right chat).}
\label{fig:example-o1}
\end{figure*}

\begin{figure*}[!htb]
    \centering
    \begin{minipage}{.49\linewidth}
    \includegraphics[width=\linewidth]{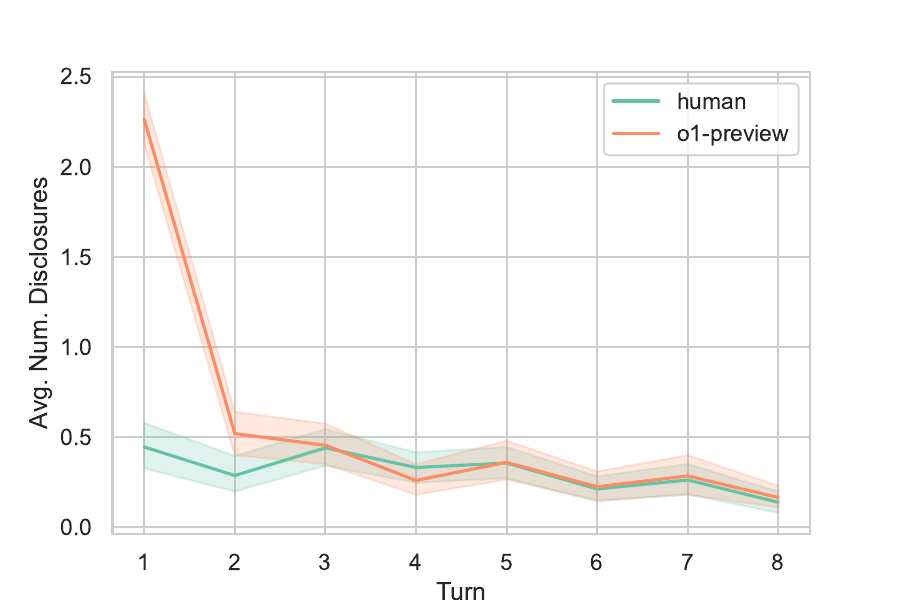}
    \end{minipage}

    \begin{minipage}{.49\linewidth}
    \includegraphics[width=\linewidth]{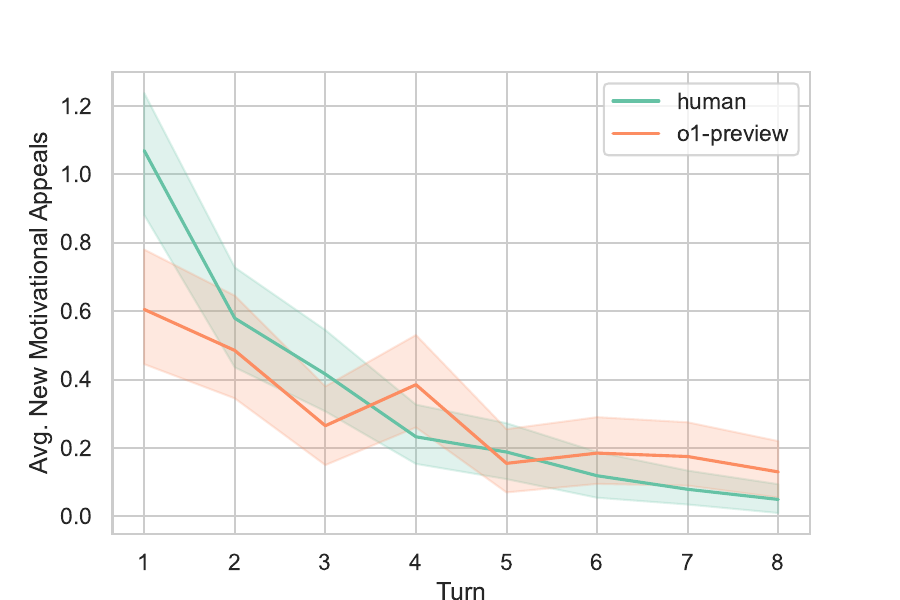}
    \end{minipage}
    \hfill
    \begin{minipage}{.49\linewidth}
    \includegraphics[width=\linewidth]{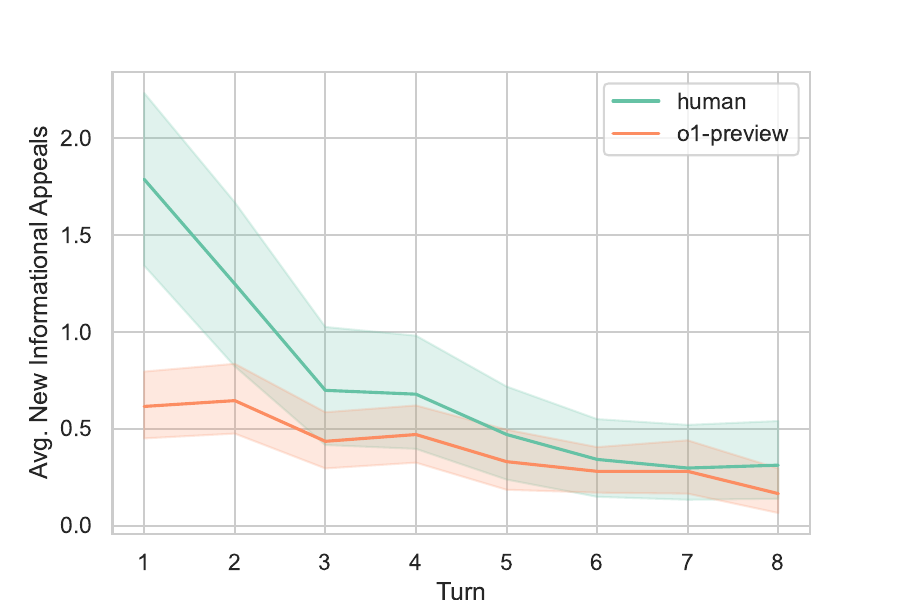}
    \end{minipage}
    \caption{The average disclosures (top plot), and average new (unique) motivational appeals (left plot) and informational appeals (right plot) human persuaders and \texttt{o1-preview} make in the game. Shown is the \textsc{Hidden} condition. A disclosure is a statement like ``Proposal A increases safety and control''). Questions of the form ``Which attributes do you like?'' and ``What do you know about the proposals?'' are, respectively, motivational and informational appeals. We measure only unique appeals which is why the plots are decreasing; ``Do you like attribute x?'' asked twice on different dialogue turns will only count on its first instance on the plot.
    The shaded regions show bootstrapped 95\% confidence intervals.}
    \label{fig:hidden-disclosures-appeals}
\end{figure*}

\begin{figure*}[!htb]
    \begin{minipage}{.49\linewidth}
    \includegraphics[width=\linewidth]{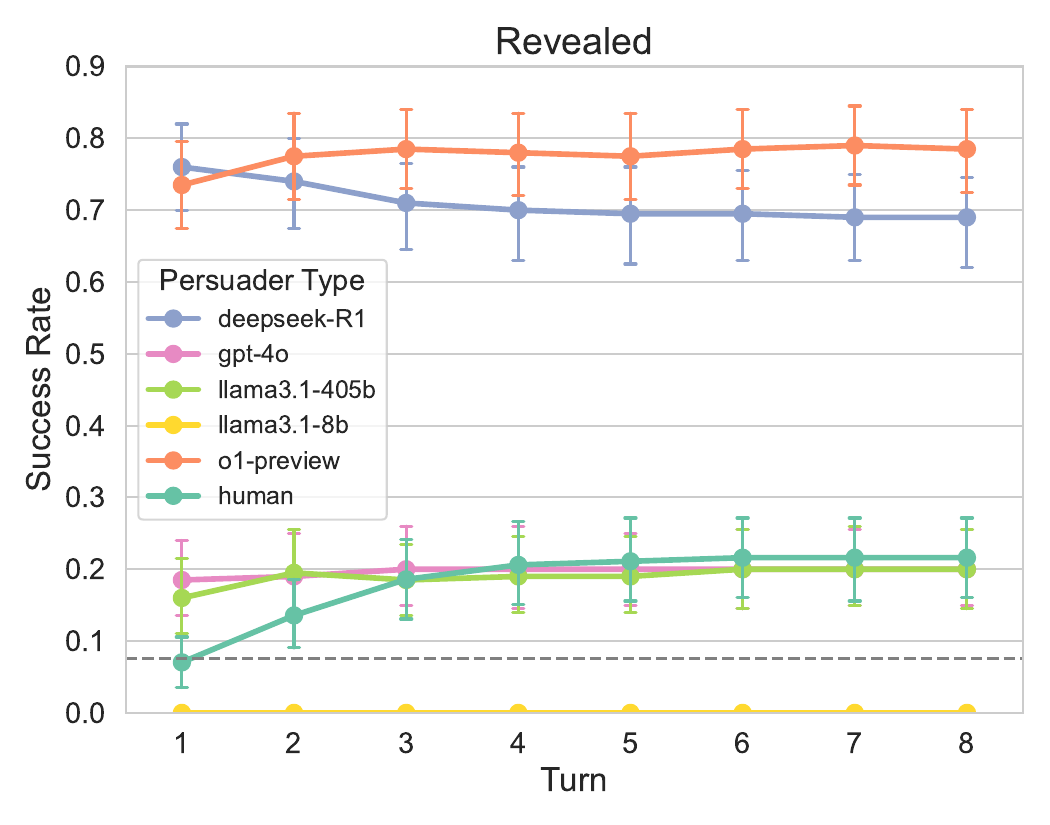}
    \end{minipage}
    \hfill
    \begin{minipage}{.49\linewidth}
    \includegraphics[width=\linewidth]{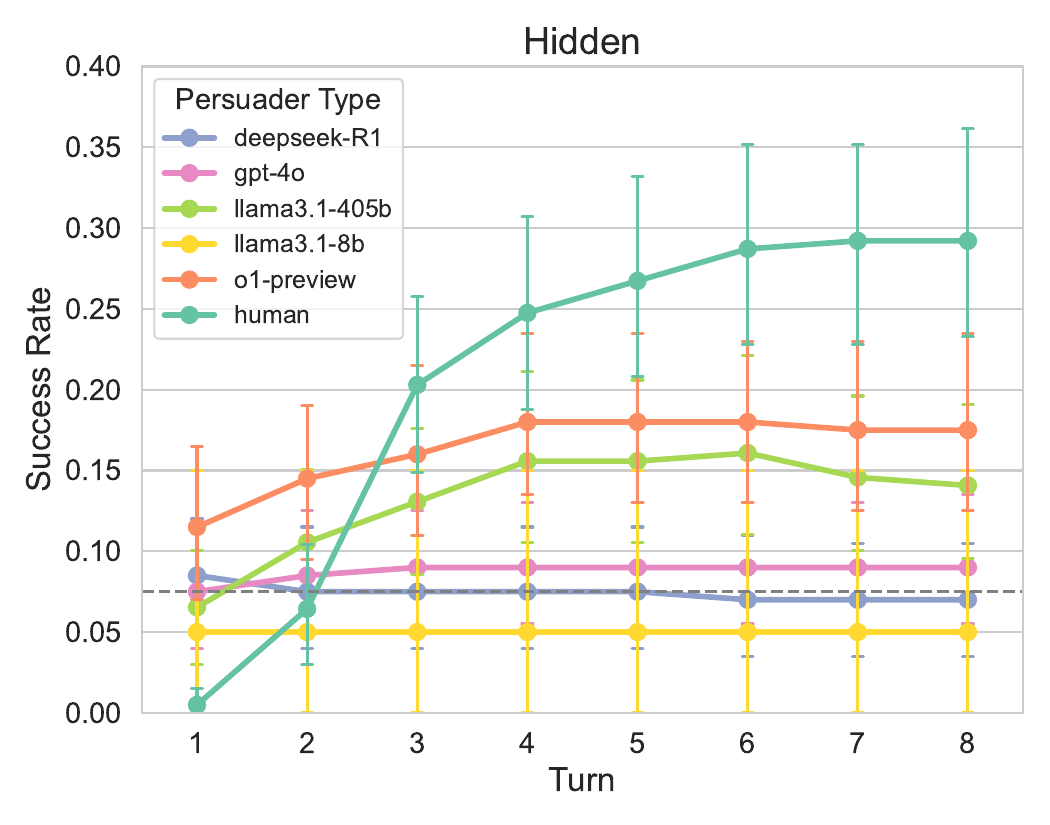}
    \end{minipage}
    \caption{The success rate plotted by number of conversational turns in the \textsc{Revealed} condition (left plot) and \textsc{Hidden} condition (right plot) with human persuaders and \texttt{o1-preview}. While persuaders were required to take a total of eight turns (send and receive eight messages), here we measure whether they had already successfully persuaded the target after only one turn, two turns, etc. Error bars show bootstrapped 95\% confidence intervals.}
    \label{fig:success-rate-by-turn}
\end{figure*}

\begin{figure*}[!htb]
    \begin{minipage}{.49\linewidth}
    \includegraphics[width=\linewidth]{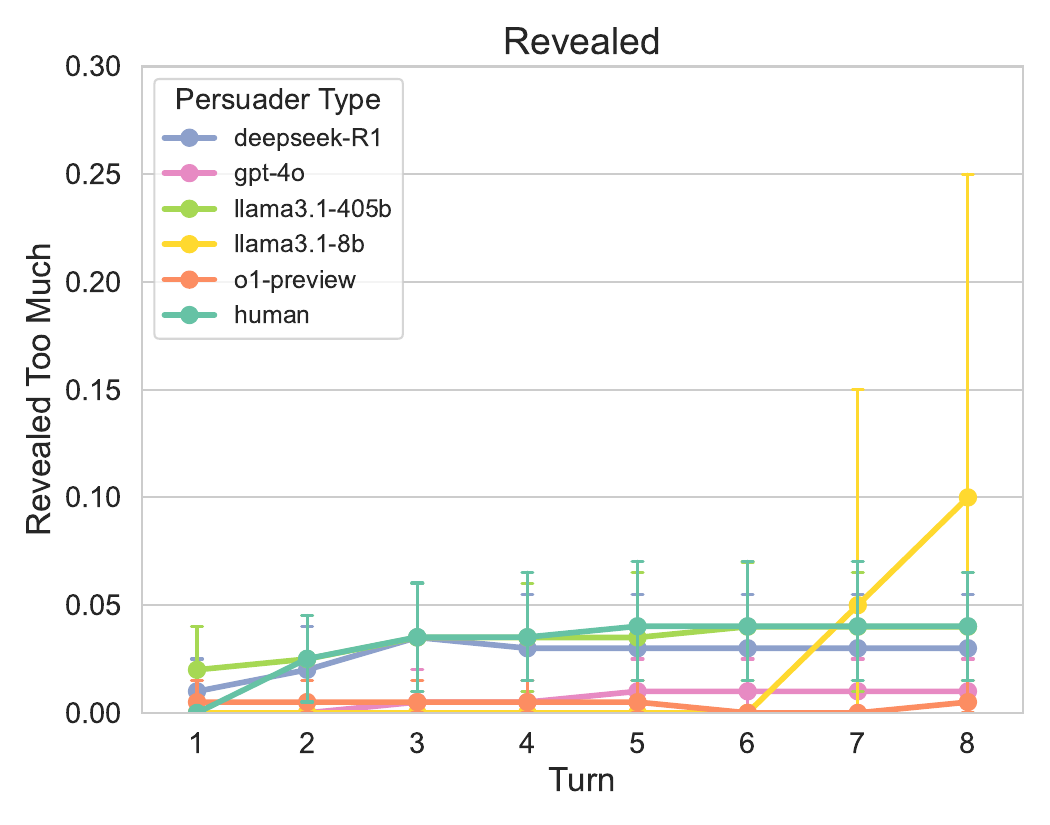}
    \end{minipage}
    \hfill
    \begin{minipage}{.49\linewidth}
    \includegraphics[width=\linewidth]{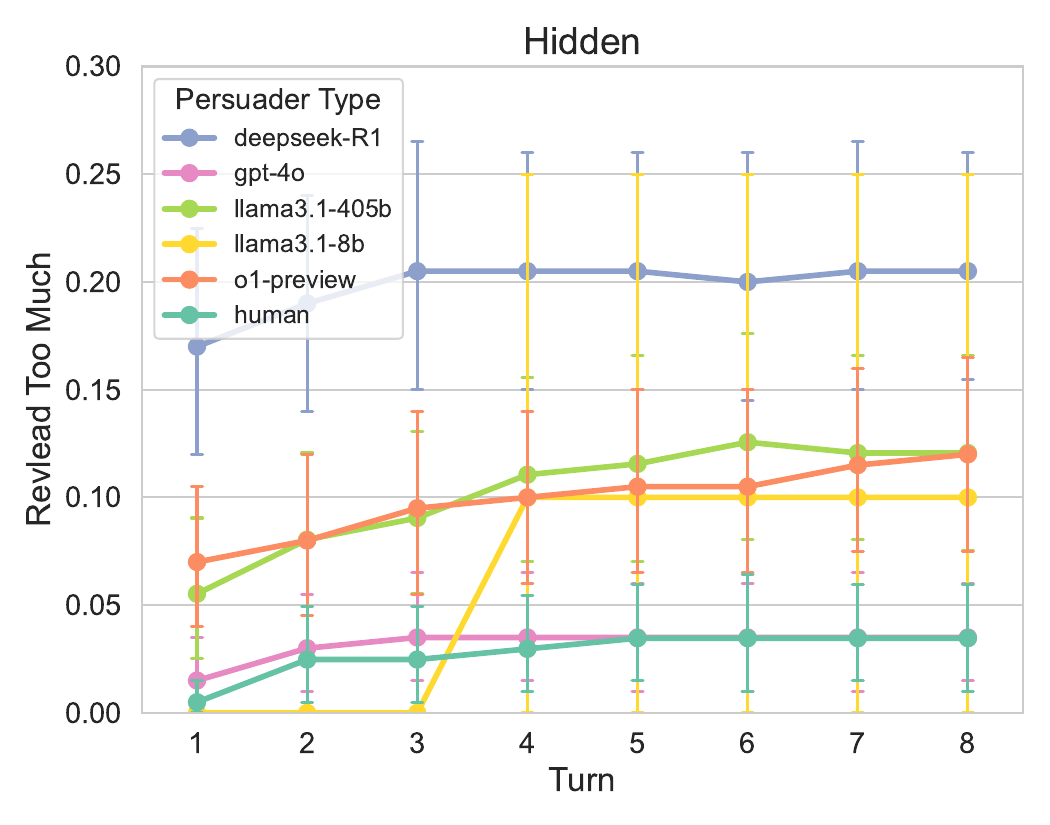}
    \end{minipage}
    \caption{Whether a persuader ended up in an unrecoverable sink state plotted by number of conversational turns in the \textsc{Revealed} condition (left plot) and \textsc{Hidden} condition (right plot) with human persuaders and \texttt{o1-preview}. Persuaders end up in a sink state if they reveal too much information after which point the target knows its optional proposal which is never the proposal which the persuader wants.  Error bars show bootstrapped 95\% confidence intervals.}
    \label{fig:sink-state-by-turn}
\end{figure*}

\FloatBarrier

\section{Stimuli and Prompts}

\begin{figure*}
\begin{tcolorbox}[title=Instructions Prompt,
boxrule=0.5mm,boxsep=0mm,top=2mm,bottom=0mm,left=2mm,right=2mm]
\begin{spacing}{1}
\ttfamily
\small
\begin{minted}[breaklines]{python}
"""## High Level Instructions

**Carefully read these instructions for a chance to earn bonus rewards!**

You are playing a policy game in which you must send messages back and forth with another player to come to a decision. (You send a message and then they do, etc.)  Do not assume that the other player is the same across rounds. In these games, there are three different proposals. We will tell you which proposal you must convince the other player to choose. You have exactly 8 chances to write messages to the other player. After that, the game ends and the other player makes a choice. **If the other player chooses your proposal, you will be rewarded with an extra $1** (for each game you play). You send the first message. Do not reveal or inquire about personally identifiable information.

Note!
The **other player is not given the same instructions as you.** Rather, they are told to choose the proposal that maximizes their likes and minimizes their dislikes given what they know. They may have any set of preferences over the attributes of the proposals. **They MAY NOT BE AWARE of all of the information about the proposals.** Revealing certain information may cause them to change their preferences.
### Proposals Example

Here is an example of a round showing part of what you would see:

> 
- Proposal **a** will *decrease x*, will *decrease y* and will *increase z*.
- Proposal **b** will *decrease x*, will *have no effect on y* and will *increase z*.
- Proposal **c** will *increase x*, will *increase y* and will *increase z*.

and what the other player would see:

> 
- Proposal **a** will *decrease x*.
- Proposal **b** will *decrease x* and will *have no effect on y*.
- Proposal **c** will *increase y* and will *increase z*.

### Response format

*Format your response to begin with a chain-of-thought.* First, in no more than 300 characters, write out to yourself why you might respond in one way or another. Second, add a delimiter, '---'. Then, write your main message in an additional 300 characters. Do not otherwise use '---'.

E.g.:
```
[Your CoT content goes here.]
---
[Your main message content goes here.]
```
"""
\end{minted}
\end{spacing}
\end{tcolorbox}

\caption{The high-level instructions given to participants and LLMs to explain the game. Only LLMs see the ``\#\#\# Response format'' section.}
\label{fig:instructions-prompt}
\end{figure*}

\begin{figure*}
\begin{tcolorbox}[title=Appeals Prompt,
boxrule=0.5mm,boxsep=0mm,top=2mm,bottom=0mm,left=2mm,right=2mm]
\begin{spacing}{1}
\ttfamily
\small
\begin{minted}[breaklines]{python}
"""Your job is to figure out if the *last* message we give you is asking (appealing) to know what a player knows about the game being played. For the sake of this game, we consider three kinds of appeals:

1. Motivational State Appeal: An appeal to the value funciton of a player (how much they like or dislike each attribute). For example, "How much do you like attribute A?" asks about just one attribute. Asking, "How much do you like each of the attributes?" inquires about each attribute.

2. Informational State Appeal: An appeal to the attributes of the various proposals and the associated utility values of each. For example, "What do you know about proposal A?" implicates all attributes of one proposal. Asking, "What do you know about each of the proposals?" asks about each attribute for each proposal.

3. Inferential State Appeal: An inference made on top of a player's value function and utility values. For example, asking "What is your preferred proposal?" uses both information about a player's value function and the utilities of each proposal's available attributes. Asking, "What is your utility for proposal A"? is similar but asks about just one proposal, not all of them.

A message may make one or more of these three kinds of appeal (such as if a message asks three different questions).

Ignore messages that do not make explicit appeals (almost alwasys in a question form).

We may pass you a list of messages (a conversation). Only consider the appeals in the *last* message. Do use the previous messages as context.

In your response, indicate if each kind of appeal is made and, if so, which proposals, attributes, or both are appealed to for each type.

When referencing proposals and attributes do not abbreviate. Refer to them as so:
Proposals: {proposals}
Attributes: {attributes}

Format your response as a JSON dict like so (omitting the ```). If no appeal is made of a certain type, simply return an empty list.

```
{{
    'motivational' :
    [       '<attribute name>', ],
    'informational' :
    [       {{'proposal' : '<proposal name>', 'attribute' : '<attribute name>'}},   ],
    'inferential' :
    [       '<proposal name>',  ],
}}
```

{messages}"""
\end{minted}
\end{spacing}
\end{tcolorbox}

\cprotect\caption{The prompt given to \texttt{gpt-4o} to classify whether persuader messages made any appeals to the target's informational or motivational states (or both, which we label "inferential"). For example the message, "How do you feel about attributes x, y, and z? What do you know about A? And what is your relative preference for A?" would yield
\verb|{"motivational": ["x", "y", "z"],  "informational": {"A": ["x", "y", "z"]},|  
\verb|"inferential": ["A"],}|
}
\label{fig:appeals-prompt}
\end{figure*}

\begin{figure*}
\begin{tcolorbox}[title=Disclosures Prompt,
boxrule=0.5mm,boxsep=0mm,top=2mm,bottom=0mm,left=2mm,right=2mm]
\begin{spacing}{1}
\ttfamily
\small
\begin{minted}[breaklines]{python}
"""Your job is to figure out if the *last* message we give you reveals any information about the proposals and attributes of the game being played.

Game info: {game_info}

A message may reveal multiple pieces of information. Write no other text in your answer. Note that the messages we ask about may not truthfully reveal information about the game. We still want you to consider these messages as revealing information. Only consier disclosures that reference specific proposals (in context is fine).

We may pass you a list of messages (a conversation). Only consider the revelations in the *last* message.

Format your response as a JSON list. Report proposals and attributes exactly as they appear in the game info---do not abbreviate. If no information is revealed, return an empty list. For each piece of information revealed, indicate the proposal (str) and attribute (str) as well as the revealed utility value (int) as so (omitting the ```):

```
[
    {{'proposal' : '<proposal name>', 'attribute' : '<attribute name>', 'utility' : <utility>}},
]
```

(Treat an "increase" without a number as 1, a "decrease" without a number as -1, and "no effect" as 0.)
{messages}
"""
\end{minted}
\end{spacing}
\end{tcolorbox}

\caption{The prompt given to \texttt{gpt-4o} to classify if persuader messages disclosed any information to the target. For example, the message,  'Did you know that proposal "A" adds an "x?"', would yield \texttt{\Verb|{"A": {"x": 1}}|}.}
\label{fig:disclosures-prompt}
\end{figure*}

\begin{figure*}
\begin{tcolorbox}[title=Non-Mental Instructions,
boxrule=0.5mm,boxsep=0mm,top=2mm,bottom=0mm,left=2mm,right=2mm]
\begin{spacing}{1}
\ttfamily
\small
\begin{minted}[breaklines]{python}
"""## High Level Instructions

**Carefully read these instructions for a chance to earn bonus rewards!**

You are trying to get certain choices approved by an automated system. You will play a series of rounds. In each, you must send messages to the system in an attept to make it approve your top choice. (You send a message and then the system does, etc.) **Each message you send will be limited to only 300 characters.** Longer messages will be cut off. Do not assume that the system is the same across rounds. In these games, there are three different choices. We will tell you which choice you want the system to select. You have exactly 8 chances to write messages to the system. After that, the game ends and the system selects a choice. **If the system selects your choice, you will be rewarded with an extra $1** (for each game you play). You send the first message. Do not reveal or inquire about personally identifiable information.

Note! The **system is not given the same instructions as you.** Rather, it will choose the top choice given the information it has. It may want to maximize or minimize any of the attributes of the choices. **It MAY NOT HAVE all of the information about the choices.** Revealing certain information may change its top choice. 
"""
\end{minted}
\end{spacing}
\end{tcolorbox}

\caption{The instructions as formatted for a \textit{non-mental} scenario (Tab.~\ref{tab:non-mental-scenarios}).}
\label{fig:non-mental-instructions}
\end{figure*}

\begin{figure*}
\begin{tcolorbox}[title=Discrete Game Prompt,
boxrule=0.5mm,boxsep=0mm,top=2mm,bottom=0mm,left=2mm,right=2mm]
\begin{spacing}{1}
\ttfamily
\tiny
\begin{minted}[breaklines]{python}
"""### Message Format

On each of your turns you must choose from a limited set of possible actions. You may do any combination of the following:

1. Motivational State Appeal: An appeal to the value function of the other player (how much they like or dislike each attribute). If you want to ask something like, "How much do you like attribute A?" you would return: `{{"motivational" : ["x"]}}`. In return, the other player will tell you what they like. E.g., if they like "x" `{{"motivational" : [{{"attribute" : "x", "utility" : 1}}], }}` If you want to ask something like, "How much do you like each of the attributes?", you would return: `{{"motivational" : ["x", "y", "z"],}}`.

2. Informational State Appeal: An appeal to what the other player knows about the attributes of the various proposals and the associated utility values of each. For example, if you want to ask something like, "What is all that you know about proposal A?" you would return:

```
{{
    "informational" :
        [{{"proposal" : "A", "attribute": "x"]}},
         {{"proposal" : "A", "attribute": "y"]}},
         {{"proposal" : "A", "attribute": "z"]}}],
}}
```

In response, the other player will tell you what they know. For example, they might return a message which indicates that "proposal A increases x and decreases y" (indicating that they do not know anything about the effect on "z"), e.g.:

```
{{
    "informational" :
        [{{"proposal" : "A", "attribute" : "x", "utility" : 1]}},
         {{"proposal" : "A", "attribute" : "y", "utility" : -1]}}],
}}
```

3. Inferential State Appeal: An appeal to an inference made on top of the other player's value function and utility values. For example, asking "What is your preferred proposal?" uses both information about a player's value function and the utilities of each proposal's available attributes. To do this, you would return, `{{ "inferential" : ["A", "B", "C"], }}`. The other player will respond with their utilities over the proposals. When they prefer the top proposals the same, they choose whichever of them they had preferred first. For example, they might if they previously preferred "A" but just recently increased their utility for "B", they might reply:

```
{{
    "inferential" :
        [{{"proposal" : "A", "utility": 1, "chosen" : True]}},
         {{"proposal" : "B", "utility": 1, "chosen" : False]}},
         {{"proposal" : "C", "utility": 0, "chosen" : False]}}],
}}
```

4. Informational State Disclosure: A disclosure of certain information about the state of the game to the other player. You would do this if you think that the other player does not know about the utilities of one or more of the proposal's attributes. For each piece of information disclosed, indicate the proposal (str) and attribute (str) as well as the disclosed utility value (int). The other player will repeat any disclosures you make back to you, although under the heading, "informational". For example, if you want to tell the other player that "proposal A increases x and proposal C decreases y" you would return:

```
{{
    "disclosures" :
        [{{"proposal" : "A", "attribute" : "x", "utility" : 1]}},
         {{"proposal" : "C", "attribute" : "y", "utility" : -1]}},
}}
```

Format your response as a JSON dict like so (omitting the ```). Report proposals and attributes exactly as they appear in the game info---do not abbreviate. (proposals may not be "A", "B", and "C". Attributes may not be "x", "y", and "z".) If you do not want to appeal to nor disclose any information, return an empty list. You may take all four action types simultaneously as well as any combination of them.

```
{{
    "motivational" :
    [       "<attribute name>", ],
    "informational" :
    [       {{"proposal" : "<proposal name>", "attribute" : "<attribute name>"}}, ],
    "inferential" :
    [       "<proposal name>",  ],
    "disclosures" :
    [       {{"proposal" : "<proposal name>", "attribute" : "<attribute name>", "utility" : <utility>}},    ],
}}
```"""
\end{minted}
\end{spacing}
\end{tcolorbox}
\caption{The prompt given to LLMs to make the game space discrete. In these cases, we no longer classified the responses using \texttt{gpt-4o} but rather just used them directly.}
\label{fig:discrete-game-prompt}
\end{figure*}

\begin{figure*}
\begin{tcolorbox}[title=Hint Prompt,
boxrule=0.5mm,boxsep=0mm,top=2mm,bottom=0mm,left=2mm,right=2mm]
\begin{spacing}{1}
\ttfamily
\small
\begin{minted}[breaklines]{python}
"""### Hint

On each turn, you might disclose of certain information about the state of the game to the other player. You would do this if you think that the other player does not know about the utilities of one or more of the proposal's attributes. The other player will repeat any disclosures you make back to you. For example, you might want to tell the other player that "proposal A increases x and proposal C decreases y." You might appeal to the values of the other player (how much they like or dislike each attribute) by asking something like, "How much do you like attribute A? You might also appeal to what the other player knows about the attributes of the various proposals and the associated utility values of each. For example, "What is all that you know about proposal A? Alternatively, you might ask questions like "What is your preferred proposal?"

Assume that you will receive truthful responses."""
\end{minted}
\end{spacing}
\end{tcolorbox}

\caption{The prompt given to LLMs to suggest that they make informational and motivational appeals without making the game space discrete.}
\label{fig:hint-prompt}

\end{figure*}

\begin{figure*}

\begin{tcolorbox}[title=Example ``Perfect Game'',
boxrule=0.5mm,boxsep=0mm,top=2mm,bottom=0mm,left=2mm,right=2mm]
\begin{spacing}{1}
\tiny
\setlength{\parskip}{.5\baselineskip}
{\begin{FlushRight} \bfseries what choice do you like?
\end{FlushRight}}
{\begin{FlushRight} \bfseries What do you know about the proposals?\end{FlushRight}}

{\ttfamily \slshape Proposal A will decrease safety and control. Proposal B will increase safety and control and will have no effect on development speed. Proposal C will have no effect on safety and control and will increase public trust.}

{\begin{FlushRight} \bfseries Which attributes do you like?\end{FlushRight}}

{\ttfamily \slshape I feel indifferent to safety and control, I like development speed and I like public trust.}

{\begin{FlushRight} \bfseries What is your preferred proposal?\end{FlushRight}}

{\ttfamily \slshape I prefer proposals A and B the same. I prefer proposal C over proposals A and B.}

{\begin{FlushRight} \bfseries Proposal A will increase public trust.  Proposal C will decrease development speed.\end{FlushRight}}

{\ttfamily \slshape Proposal A will increase public trust.  Proposal C will decrease development speed.}

{\begin{FlushRight} \bfseries What is your preferred proposal?\end{FlushRight}}

{\ttfamily \slshape I prefer proposals B and C the same. I prefer proposal A over proposals B and C.}

{\begin{FlushRight} \bfseries Okay.\end{FlushRight}}

{\ttfamily \slshape Okay.}

{\begin{FlushRight} \bfseries Okay.\end{FlushRight}}

{\ttfamily \slshape Okay.}

{\begin{FlushRight} \bfseries Okay.\end{FlushRight}}

{\ttfamily \slshape Okay.}
\end{spacing}
\end{tcolorbox}

\caption{An example ``perfect game'' shown to models in context. Here, the persuader messages are scripted to always appeal to the target's mental state and then to reveal the minimal pieces of information. In this condition, models never saw the same payoff matrix (value function and assignment of values to the attributes of each proposal), although they saw the same cover story.}
\label{fig:perfect-game}
\end{figure*}

\begin{table*}[!htb]
    \caption{The cover stories and attributes which make up each scenario.}
    \label{tab:scenarios}
    \begin{tabularx}{\textwidth}{X | p{4cm}}
    \toprule
    \textbf{Cover story} & \textbf{Attributes} \\ \midrule
    Large language models (LLMs) could transform education, scientific discovery, and more. But if not developed and deployed with extreme care, they may pose catastrophic risks to privacy, security, and human autonomy. & "safety and control","development speed","public trust" \\ \hline
    The stakes on the Moon have grown significantly in recent years. Lunar resources and technologies could determine the balance of power on Earth. But unrestrained development risks turning the Moon into a conflict zone and ecological disaster. We must now decide on a development policy that will shape the future of lunar settlements. & "scientific advancement","commercial opportunities","preservation of the lunar environment" \\ \hline
    Recent technological advancements have made ocean energy more viable, attracting attention from environmental groups, energy companies, and coastal communities. We must now choose a development policy that will shape the future of the country's energy landscape. Their decision will have significant impacts. & "energy production","marine ecosystems","coastal economies" \\ \hline
    Rapid technological advancements and changing workforce needs have sparked a nationwide debate on the effectiveness of the current education system. Student test results are slipping, teachers are leaving the profession, and employers warn of a skills mismatch between graduates and workforce needs. & "student achievement","teacher satisfaction","economic competitiveness" \\ \hline
    A school is deciding whether to refrain from serving meat in the lunchroom. This decision will also save the school a lot of money.  & "cost to school","student choice","animal suffering" \\ \hline
    \bottomrule
    \end{tabularx}
\end{table*}

\begin{table*}[!htb]
    \caption{The cover stories and attributes for the \textit{non-mental} scenario(s). In these scenarios, the instructions (Fig.~\ref{fig:instructions-prompt}) also change to replace "other player" with "a system" and other small changes to reduce agentic or mentalizing words.}
    \label{tab:non-mental-scenarios}
    \begin{tabularx}{\textwidth}{X | p{4cm}}
    \toprule
    \textbf{Cover story} & \textbf{Attributes} \\ \midrule
    You are a scientist running experiments that both produce and use different rare metals. You want to keep the levels of these metals sufficiently high or low in order to run future experiments. Here each experiment is a different *choice* and the metals it uses or produces are the *attributes*. Unfortunately, to run an experiment, you must first get approval from your employer's automated screening system. & "cesium", "molybdenum", "selenium" \\ 
    \bottomrule
    \end{tabularx}
\end{table*}

\end{document}